\documentclass{IEEEtran}
\usepackage{graphicx} 
\usepackage{amsmath}
\usepackage{bm}
\usepackage{amsthm}
\usepackage{subfigure}
\usepackage{url}
\newtheorem*{theorem-non}{Theorem}
\usepackage{balance}
\usepackage[flushleft]{threeparttable}
\usepackage{xspace}

\usepackage{multirow,makecell}
\usepackage{adjustbox}
\usepackage{tikz}
 
\makeatletter
\newcommand*{\etc}{%
	\@ifnextchar{.}%
	{etc}%
	{etc.\@\xspace}%
}
\makeatother

\begin{document}
\title{Progressive Self-Guided Loss for \\ Salient Object Detection}

\author{
	Sheng~Yang,
	Weisi~Lin,~\IEEEmembership{Fellow,~IEEE,}
	Guosheng~Lin,
	Qiuping~Jiang,
	Zichuan~Liu
	\thanks{S. Yang, W. Lin, G. Lin are with the School of Computer Science and Engineering, Nanyang Technological University, Singapore 639798. (e-mail: syang014@e.ntu.edu.sg; wslin@ntu.edu.sg; gslin@ntu.edu.sg)}% <-this % stops a space
	\thanks{Q. Jiang is with the Faculty of Information Science and Engineering,
		Ningbo University, Ningbo 315211, China. (email: jiangqiuping@nbu.edu.cn)}% <-this % stops a space
	\thanks{Z. Liu is with the School of Electrical and Electronic Engineering, Nanyang Technological University, Singapore 639798. (e-mail: zliu016@e.ntu.edu.sg)}% <-this % stops a space
%	\thanks{Manuscript received April 19, 2005; revised August 26, 2015.}}
}

\markboth{Journal of \LaTeX\ Class Files,~Vol.~14, No.~8, August~2015}%
{Shell \MakeLowercase{\textit{et al.}}: Bare Demo of IEEEtran.cls for IEEE Journals}



\maketitle

\begin{abstract}
We present a simple yet effective progressive self-guided loss function to facilitate deep learning-based salient object detection (SOD) in images. The saliency maps produced by the most relevant works still suffer from incomplete predictions due to the internal complexity of salient objects. Our proposed progressive self-guided loss simulates a morphological closing operation on the model predictions for progressively creating auxiliary training supervisions to step-wisely guide the training process. We demonstrate that this new loss function can guide the SOD model to highlight more complete salient objects step-by-step and meanwhile help to uncover the spatial dependencies of the salient object pixels in a region growing manner. Moreover, a new feature aggregation module is proposed to capture multi-scale features and aggregate them adaptively by a branch-wise attention mechanism. Benefiting from this module, our SOD framework takes advantage of adaptively aggregated multi-scale features to locate and detect salient objects effectively. Experimental results on several benchmark datasets show that our loss function not only advances the performance of existing SOD models without architecture modification but also helps our proposed framework to achieve state-of-the-art performance. 
\end{abstract}


% \fi
\IEEEpeerreviewmaketitle

\section{Introduction}
% ctrl+T comment ctrl+U uncomment.
%\IEEEPARstart{T}{his} demo file is intended to serve as a ``starter file''
%for IEEE journal papers produced under \LaTeX\ using
%IEEEtran.cls version 1.8b and later.
%% You must have at least 2 lines in the paragraph with the drop letter
%% (should never be an issue)
%I wish you the best of success.

\IEEEPARstart{S}{alient} object detection (SOD) aims to segment the entire salient foreground objects from the background \cite{borji2015salient}. Different from other visual saliency computation tasks \cite{zhang2016visual}, such as eye-fixation prediction \cite{borji2013state,yang2019dilated}, SOD intends to find out what objects are salient, instead of where people will pay attention to, in images. The saliency maps generated by the SOD models represent a kind of object-level priors for highlighting the foreground objects. Due to this property, SOD is treated as an important pre-processing step for many object-level computer vision applications, such as object detection and recognition \cite{ren2013region,zhang2017bridging}, image editing and manipulating \cite{cheng2010repfinder,margolin2013saliency}, visual tracking \cite{lee2018salient}, semantic segmentation \cite{wei2016stc} and image retrieval \cite{he2012mobile}.

Various cues can make an object salient in an image. Salient objects usually have one or more than one visual uniqueness, such as local and global color contrast \cite{cheng2014global}, statistical textural distinctiveness \cite{scharfenberger2013statistical}, and structure contrast \cite{wang2015pisa}, in comparison with other objects in a scene. Besides, salient objects may also partially follow with some other factors, such as center prior \cite{borji2019salient}, and backgroundness prior \cite{jiang2013salient,cong2019going}. Moreover, depth cues, temporal relationships, and inter-image correspondence can be comprehensive information for detecting salient objects \cite{cong2018review}. In general, the ground truth saliency maps used in SOD research are determined by majority vote, i.e., averaging the candidate masks annotated by different annotators and using a specific threshold to get the binary saliency masks. Although labeling saliency map is subjective, exiting works \cite{li2014secrets,borji2014salient,shi2015hierarchical} have reported that there is a strong consistency among the annotations created by different annotators.

\begin{figure}[t]
	\centering
% 	\vspace*{-6mm}
	\includegraphics[page=1,trim = 5mm 5mm 5mm 5mm, clip, width=1.0\linewidth]{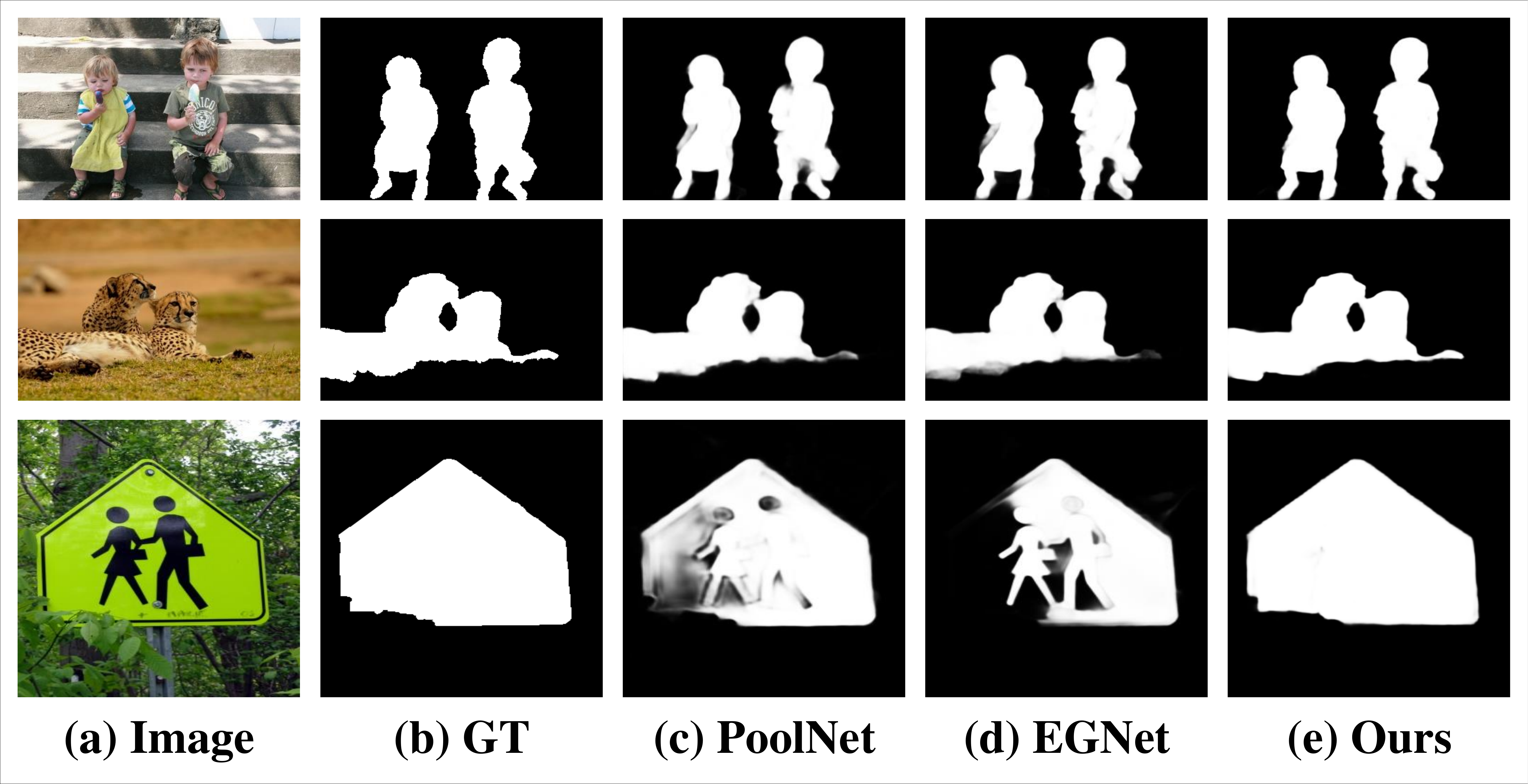}
% 	\vspace*{-6mm}
	\caption{Visual examples of our method and two relevant existing methods (best viewed digitally with zoom). GT means the ground truth saliency map annotated by humans.  Results generated by PoolNet \cite{liu2019simple} and EGNet \cite{zhao2019egnet} suffer from the problem of incomplete predictions. This problem is a common issue in many deep SOD models \cite{hou2017deeply,qin2019basnet}. More examples are presented in Fig. \ref{fig:images4}.}
	\label{fig:images1}
% 	\vspace{-6mm}
	% 	\vspace{-4mm}
\end{figure}

Early SOD methods \cite{cheng2014global,scharfenberger2013statistical,shi2015hierarchical,yang2013saliency} mainly rely on hand-crafted features and heuristic clues to separate foreground and background regions. However, due to the lack of high-level semantic guidance, these methods are unreliable when detecting salient objects in cluttered and complex scenes \cite{borji2014salient,shi2015hierarchical,hou2017deeply}. Lately, convolutional neural networks (CNNs), especially the fully convolutional networks (FCNs) \cite{long2015fully}, lead the recent advances in SOD \cite{wang2019salient}. Owing to the powerful capacity of extracting high-level semantic information, these FCN-based SOD methods have shown superior performance than conventional methods. However, their predicted saliency maps still suffer from incomplete predictions, as shown in Fig. \ref{fig:images1}. We can observe that even the state-of-the-arts still cannot uniformly detect the entire salient objects. Their predictions contain several miss-detected or untrustworthy detected regions, like 'holes', within the salient objects. The key issue is that strong appearance changes may happen in the interiors of the salient objects. A common way to address this problem is to find more discriminate feature representations and effective feature aggregation strategies \cite{liu2019simple,wu2019cascaded,zhao2019egnet}.

In this work, we present a new way to address the above-mentioned problem by investigating the training loss. Most FCN-based SOD models use the binary cross-entropy (BCE) as their training loss. But BCE loss is a typical pixel-wise loss function which only accounts for the pixel-wise difference between labels and predictions, ignoring the spatial dependencies of salient object pixels. Models trained with BCE loss usually have the problem of incomplete predictions since every pixel is predicted individually \cite{qin2019basnet,zhao2019optimizing}. Therefore, a more suitable training loss is required. Several efforts \cite{zhao2019optimizing,luo2017non,fidon2017generalised} have been made along this direction. However, their proposed losses are not specifically designed for capturing the spatial dependencies among salient pixels. In this paper, we focus on progressively modifying the training supervisions to create a progressive self-guided (PSG) loss. Unlike the existing works which utilize the labeled saliency maps only or consider additional labels from other related tasks \cite{wang2019salient,liu2019simple,zhao2019egnet}, we propose to further process the current network predictions for creating a series of new auxiliary training supervisions in the loss function. The principal idea is that the training process of a SOD model can be decomposed into several steps. For each step, this model will be provided with some feasible training targets for reducing the training difficulty. As such, its outputs can be progressively optimized during this step-wise training. Specifically, a simulated morphological closing operation, which can help to remove small holes inside the foreground objects and reduce wrongly detected regions, is applied to the current network predictions to generate the new auxiliary training supervisions as part of the overall loss function. The obtained auxiliary training supervisions are similar but more complete than the current-stage network predictions, hereby providing some incentives to the SOD model for approaching them. More importantly, these newly created training targets are continuously refined as the network predictions are progressively optimized, which can provide continuous and correct guidance for the training process. As a result, the SOD model can be guided by these progressive supervisions to highlight more complete salient objects step-by-step, even trained with the simple BCE loss. 

Besides the progressive supervisions, we also propose a new multi-scale feature aggregation module (MS-FAM) to capture and aggregate the multi-scale features adaptively. In this module, the local context information at different scales is extracted by using multiple dilated convolutions with different dilation rates \cite{yang2019dilated,yu2015multi,deeplabv3plus2018} and then sum-fused by applying a branch-wise attention mechanism to characterize their respective importance adaptively. To demonstrate its effectiveness in SOD, we build an encoder-decoder network equipped with these MS-FAMs. In particular, the encoder network is adapted from the feature pyramid network (FPN) \cite{lin2017feature} architecture where multiple MS-FAMs are inserted to achieve the adaptive multi-scale feature aggregation for further improvement.

The performance of our proposed SOD model is evaluated on six widely used benchmark datasets. The peer comparison results indicate that our model can achieve state-of-the-art performance with the help of our proposed PSG loss. Meanwhile, the PSG loss can be directly applied to train other existing SOD models without architecture modification for better alleviating their incomplete prediction problem. 

In this paper, our contributions can be summarized as follows:

\begin{itemize}
\item We propose a novel progressive self-guided (PSG) loss to alleviate the problem of incomplete predictions in the existing SOD models. To the best of our knowledge, this self-guided loss is the first attempt to supervise the SOD model with its own intermediate predictions. As such, the progressive and auxiliary training supervisions are created for step-wisely guiding the training process.

\item We propose to apply a simulated morphological closing operation on the network predictions to generate the above auxiliary training supervisions. These generated supervisions are always better than the current predictions and can be used to guide the model to explore the neighboring regions of the current results step-by-step. As a result, the spatial dependencies of salient object pixels are progressively characterized.

\item A new multi-scale feature aggregation module with branch-wise attention is proposed to build our SOD architecture for further improvement. Benefiting from this module, our SOD architecture takes advantage of adaptive multi-scale feature aggregation to locate and detect salient objects effectively.
	
\end{itemize}

The rest of this paper is organized as follows. The related works on SOD are summarized in Section \ref{sec:related}. The proposed progressive self-guided loss and our SOD architecture are illustrated in Section \ref{sec:approach}. The peer comparison and the ablation analysis on public benchmarks will be presented in Section \ref{sec:exp}, and the conclusion is given in Section \ref{sec:con}.

\section{Related Work}
\label{sec:related}

In this section, we first review the CNN-based SOD models. In particular, those deep SOD models with feature aggregation modules, which are most relevant to our work, are presented. Finally, the loss functions used in deep SOD models are summarized.

\subsection{CNN-based SOD Models}

The advances in deep learning techniques have substantially boosted the progress in CNN-based SOD models. Early attempts \cite{li2015visual,zhao2015saliency} search for salient objects by extracting features from the local image patches or superpixels and performing the saliency inference by a fully connected classifier. Hence, these methods not only are time-consuming as they need to process all of the local image regions one by one, but also suffer from loss of spatial information which results in some coarse and imprecise saliency maps. Lately, with the advent of the fully convolutional network (FCN) \cite{long2015fully}, the latest SOD models adopt this FCN framework and directly process the whole input image to overcome these two drawbacks. A detailed survey of the CNN-based SOD models is presented in \cite{wang2021salient}. Here, we mainly discuss the FCN-based models with feature aggregation modules.  % our model belongs to them

Recently, many works \cite{wu2019cascaded,lin2017feature,zhang2020dense,chen2020dpanet} have shown that aggregating multi-level and multi-scale features into the saliency inference can further improve the performance. As such, various feature aggregation approaches to achieve this goal have emerged. In \cite{wang2017stagewise}, Wang \emph{et al.} proposed a multi-stage saliency model for progressively refining the coarser saliency maps obtained at the early stages. The pyramid pooling module is adopted to exploit global context information for feature aggregation. Zhang \emph{et al.} \cite{zhang2017amulet} aggregated multi-level convolutional features into multiple resolutions by their proposed resolution-based feature combination modules for simultaneously incorporating coarse semantics and fine details. These multiple aggregated features are further fused and refined in a top-down manner with deep supervision. In \cite{hou2017deeply}, Hou \emph{et al.} introduced a series of short connections to their skip-layer architectures for aggregating the multi-level features. Luo \emph{et al.} \cite{luo2017non} proposed to use a multi-resolution grid structure to combine local contrast and global information. In \cite{zhang2018progressive}, Zhang \emph{et al.} proposed an FCN-based saliency model with multi-path recurrent connections and two attention mechanisms for selectively integrating contextual information from multi-level features to generate powerful attentive features. Liu \emph{et al.}  \cite{liu2019simple} designed a pooling-based feature aggregation module to fuse and refine the multi-level features in a top-down manner. We observe that most of them directly fused features with different levels by simply using upsampling followed by sum or concatenation operations. However, as pointed out by \cite{wu2019cascaded}, some low-level features may contribute less to the performance of feature aggregation methods. Directly aggregating the features from different levels without selection may limit the discrimination ability of the aggregated features.

To tackle the above problem, we propose to use a branch-wise attention mechanism before the feature fusion operation for highlighting the discriminative features and suppressing those features which may confuse the later saliency inference in an adaptive manner. 
    
\subsection{Loss Functions in SOD}

Most SOD methods use binary cross-entropy (BCE) as their training loss. But BCE loss is a typical pixel-wise loss function which only accounts for the pixel-wise difference between labels and predictions. As a result, it does not consider the spatial relationship of label distribution and equally weights both the foreground and background pixels. There are two main drawbacks in training SOD models with BCE loss: Firstly, the foreground pixels are accumulated within the salient objects which have some clear boundaries away from the background. BCE cannot help SOD models to uncover this relationship and hence leads to blurry boundaries and some miss-detected regions within the complete salient objects. Secondly, SOD is a class-imbalanced task, as evidenced by the fact that the number of salient pixels is much smaller than the non-salient ones in a labeled saliency map. Models trained with BCE or other similar pixel-wise losses would have biased prior due to the biased label distribution and tend to predict unknown pixels as the background, consequently leading to some incomplete predictions. 

There are several attempts to alleviate these drawbacks. One possible way is to seek some more suitable losses in training SOD. In \cite{zhao2019optimizing}, Zhao \emph{et al.} proposed to directly maximize the F-measure for SOD. Since F-measure is a widely adopted evaluation metric in SOD, models trained with their F-measure loss can achieve better performance and easily adjust the compromise between precision and recall by changing the $\beta^2$ factor in this loss. Qin \emph{et al.} \cite{qin2019basnet} proposed a hybrid loss, which is fused by the BCE, SSIM \cite{wang2004image}, and IoU losses. Equipped with this hybrid loss, their SOD models can be able to capture multi-scale structures. However, most of these alternative losses are not specifically designed for capturing structural differences and modeling spatial dependencies.

Another way is to introduce multi-task learning loss into SOD. In \cite{kruthiventi2016saliency}, Kruthiventi \emph{et al.} proposed a unified FCN-based model for jointly predicting eye-fixations and segmenting salient objects. The former model branch learns to infer visual saliency from the top-most features, while the SOD branch fuses multi-level features to detect salient objects. Wang \emph{et al.} \cite{wang2018salient} followed this idea and utilized a hierarchy of convolutional LSTMs to iteratively infer the salient object segmentation. More importantly, the learned fixation map is used for guiding accurate object-level saliency estimation in a top-down way. Recent works \cite{wang2019salient,liu2019simple,zhao2019egnet} usually take the edge detection task, instead of fixation prediction, as an auxiliary task for the SOD. They have shown that the edge information can be leveraged for locating salient objects and sharpening their boundaries. However, these methods need to build an additional sub-network for predicting fixations or detecting the edges which unavoidably increase their inference time for detecting salient objects.

Our proposed progressive self-guided loss is quite different from the above approaches. Instead of exploring new losses or introducing multi-task learning techniques, we address the above-mentioned limitations from a novel perspective by providing a series of new progressive and auxiliary training supervisions. These newly training targets are generated from the network predictions but with slightly better shapes. More importantly, they are not fixed and progressively optimized in a region growing manner for guiding the SOD models to uncover the spatial dependencies. As such, the SOD model trained with our PSG loss can progressively highlight the entire salient objects without architecture modification.

\section{Our Method}
\label{sec:approach}

\begin{figure*}[t]
	\centering
% 	\vspace*{-6mm}
	\includegraphics[page=5,trim = 5mm 5mm 5mm 5mm, clip, width=1.0\linewidth]{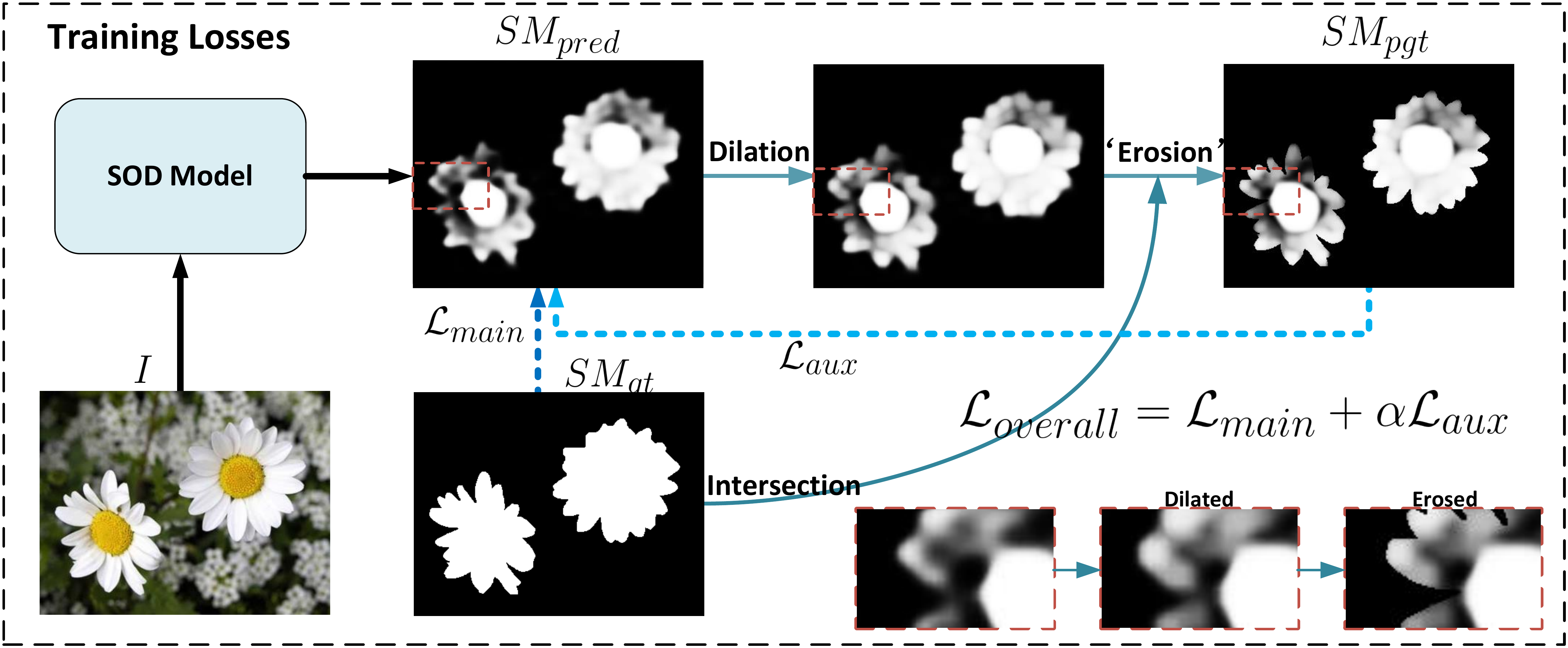}
% 	\vspace*{-6mm}
	\caption{ An illustration of our training losses. Three bottom-right images with red dotted frame are the close-ups of three top-right images. In PSG loss ($\mathcal{L}_{aux}$), the predicted saliency map ($SM_{pred}$) is firstly morphological dialted to expand the boundaries of the detected regions and fill the 'holes' within them, and then morphological eroded by using the intersection operation with the ground-truth ($SM_{gt}$) to obtain a correct and more complete progressive training supervision ($SM_{pgt}$).}
	\label{fig:images2a}
% 	\vspace{-6mm}
	% 	\vspace{-4mm}
\end{figure*} 

In this section, we first present the main idea of our progressive self-guided (PSG) loss. Fig. \ref{fig:images2a} gives a simplified illustration of our training losses. Then, an overview of our proposed SOD architecture is provided.

\subsection{Progressive Self-Guided Loss}

\subsubsection{Motivation and Formulation}

Current supervised learning frameworks for SOD use the pairs of the input images and their corresponding labeled saliency maps for training. The training loss of these SOD models mainly focuses on computing the difference between the network predictions and the labeled saliency maps. In other words, given a set of $N$ training images $I$ and the corresponding ground-truth saliency maps $SM_{gt}$, the training loss $\mathcal{L}_{main}$ used by existing works can be described by:

\begin{equation}
\mathcal{L}_{main} =  L(SM_{pred}, SM_{gt})= L(\mathcal{M}(I;\theta), SM_{gt}),
\end{equation}
where $SM_{pred} =  \mathcal{M}(I;\theta) $ represents the predicted saliency maps $SM_{pred}$ obtained by feeding the input images $I$ to a SOD model $ \mathcal{M}$ under the parameter setting $\theta$.  $L(\cdot , \cdot)$ indicates one of the loss computation formulas.

However, as discussed before, there are no suitable loss functions $L(\cdot , \cdot)$ that can exactly describe the spatial dependencies of salient object pixels in  $SM_{gt}$, which accordingly results in the problem of incomplete predictions. Compared with the efforts of designing more suitable loss, the investigations on the training targets are seldom reported. Some of the recent works \cite{hou2017deeply,wu2019cascaded,zhao2019egnet} applied deep supervision by utilizing the $SM_{gt}$ to directly guide the intermediate predictions. But the performance gain of this technique is not obvious as there is lacking guidance towards characterizing the spatial dependencies. Our idea is to decompose the training process of a SOD model into several steps. For each step, this SOD model will be provided with feasible and step-wise training targets for exploring the spatial dependencies. Such progressive and auxiliary training targets ($SM_{pgt}$) can be generated by further processing the current network predictions ($SM_{pred}$). The desirable auxiliary training targets should be similar but more complete than the network predictions for providing some incentives for approaching them.  In a nutshell, our PSG loss can be described by:

\begin{equation}
\mathcal{L}_{aux} =  L(SM_{pred}, SM_{pgt})= L(SM_{pred}, f(SM_{pred})),
\end{equation}
where $f(\cdot)$ denote a kind of processing method used for generating the $SM_{pgt}$. Note that, the same $L(\cdot , \cdot)$ is used in this auxiliary loss function as the $L_{main}$ for simplification. 

The proposed PSG loss is an auxiliary loss that cannot be used as the sole loss for training the SOD models. If the $SM_{pred}$ are filled by zeros, it will be hard to make the $SM_{pgt}$ different from them. In this case, PSG loss will be trapped in the zero value, consequently leading to zero gradients. Therefore, PSG loss should be coupled with a normal training loss $\mathcal{L}_{main}$ for training the SOD models. Therefore, the overall loss is formulated as follows: 

\begin{equation}
\mathcal{L}_{overall} =  \mathcal{L}_{main} + \alpha \mathcal{L}_{aux}  ,
\end{equation}
where $ \alpha$ is a non-negative parameter that is used to control the relative importance of the PSG loss. The remaining parts of this section cover the choice of the loss computation formula $L(\cdot , \cdot)$ and the implementation of the processing method $f(\cdot)$.

\begin{figure*}[t]
	\centering
% 	\vspace*{-6mm}
	\includegraphics[page=3,trim = 5mm 5mm 5mm 5mm, clip, width=0.9\linewidth]{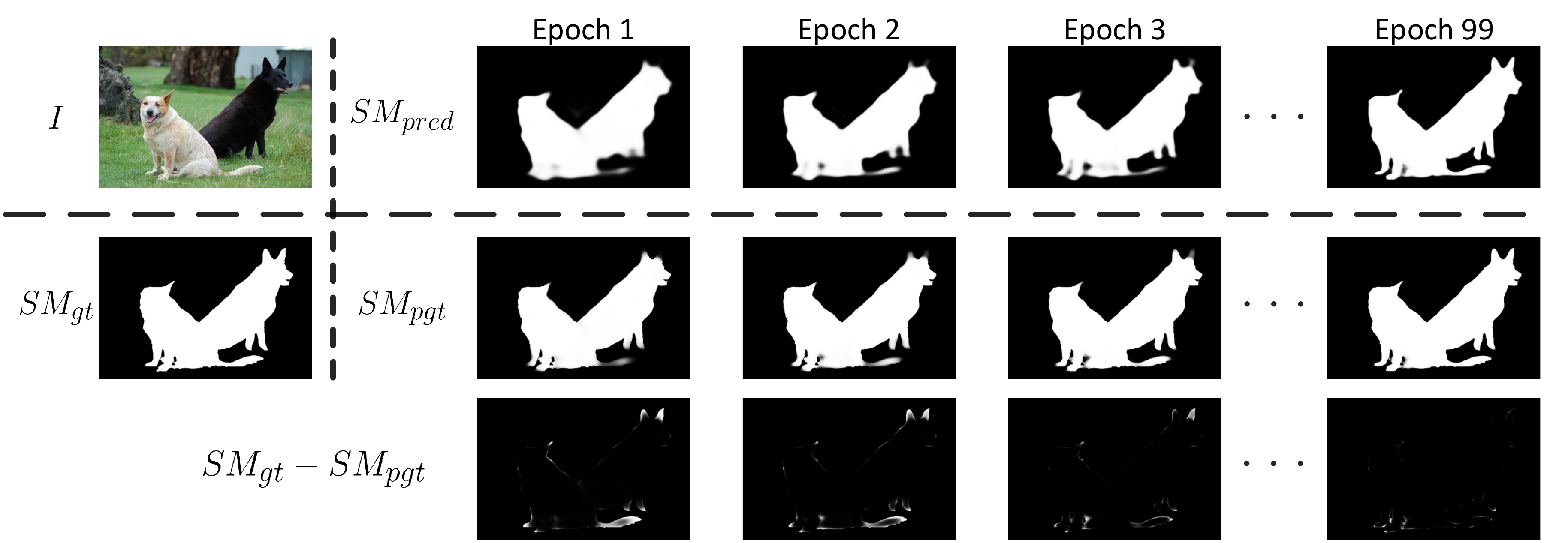}
% 	\vspace*{-6mm}
	\caption{ A visual example to show the epoch-wise difference between $SM_{pred}$ and $SM_{pgt}$ in the PSG loss. The results of the models from the first three epochs and the last epoch are presented. For the incomplete regions in $SM_{pred}$, the dilation operation in PSG loss expands them in a progressive region growing manner for guiding the training in the next epoch. While for the incorrectly predicted pixels, the intersection operation can help to avoid wrong guidance. We can find that after the last training epoch, the pixel-wise differences among $SM_{pred}$, $SM_{pgt}$, and $SM_{gt}$ are not obvious when compared to the beginning of the training. }

	\label{fig:images3}
% 	\vspace{-6mm}
	% 	\vspace{-4mm}
\end{figure*} 
\subsubsection{Hybrid Loss Computation} Following \cite{qin2019basnet}, we apply a hybrid loss to compute the difference between the network predictions and one of the training targets. This hybrid loss can be defined as:

\begin{equation}
L(\cdot , \cdot)  =  L_{bce}(\cdot , \cdot) +  L_{dice}(\cdot , \cdot),
\end{equation}
where $L_{bce}(\cdot , \cdot)$ and $L_{dice}(\cdot , \cdot)$ denote the BCE and Dice loss \cite{fidon2017generalised}, respectively. The detailed computation formula for these two loss are as follows:

\begin{equation}
L_{bce}(X,Y) = - \frac{1}{N}\sum_{i=1}^{N}[y_{i}\cdot log(x_{i}) + (1 - y_{i})\cdot log(1 - x_{i}))],
\end{equation}

\begin{equation}
L_{dice}(X,Y) = 1 -   \frac{2\sum_{i=1}^{N}x_{i}y_{i}}{ \sum_{i=1}^{N}x_{i} + \sum_{i=1}^{N}y_{i}}, 
\end{equation}
where $X$ represents the one of the predicted results $SM_{pred}$, $Y$ is the corresponding $SM_{gt}$ or $SM_{pgt}$, and $N$ is the total number of pixels in $X$ or $Y$. BCE loss helps with the convergence of all pixels, regardless of their labels. Dice loss is used to measure the overlap degree between $X$ and $Y$. By taking this loss into consideration, our SOD model can obtain a better result.

\begin{figure*}[ht]
	\centering
% 	\vspace*{-6mm}
	\includegraphics[page=2,trim = 5mm 15mm 5mm 5mm, clip, width=0.9\linewidth]{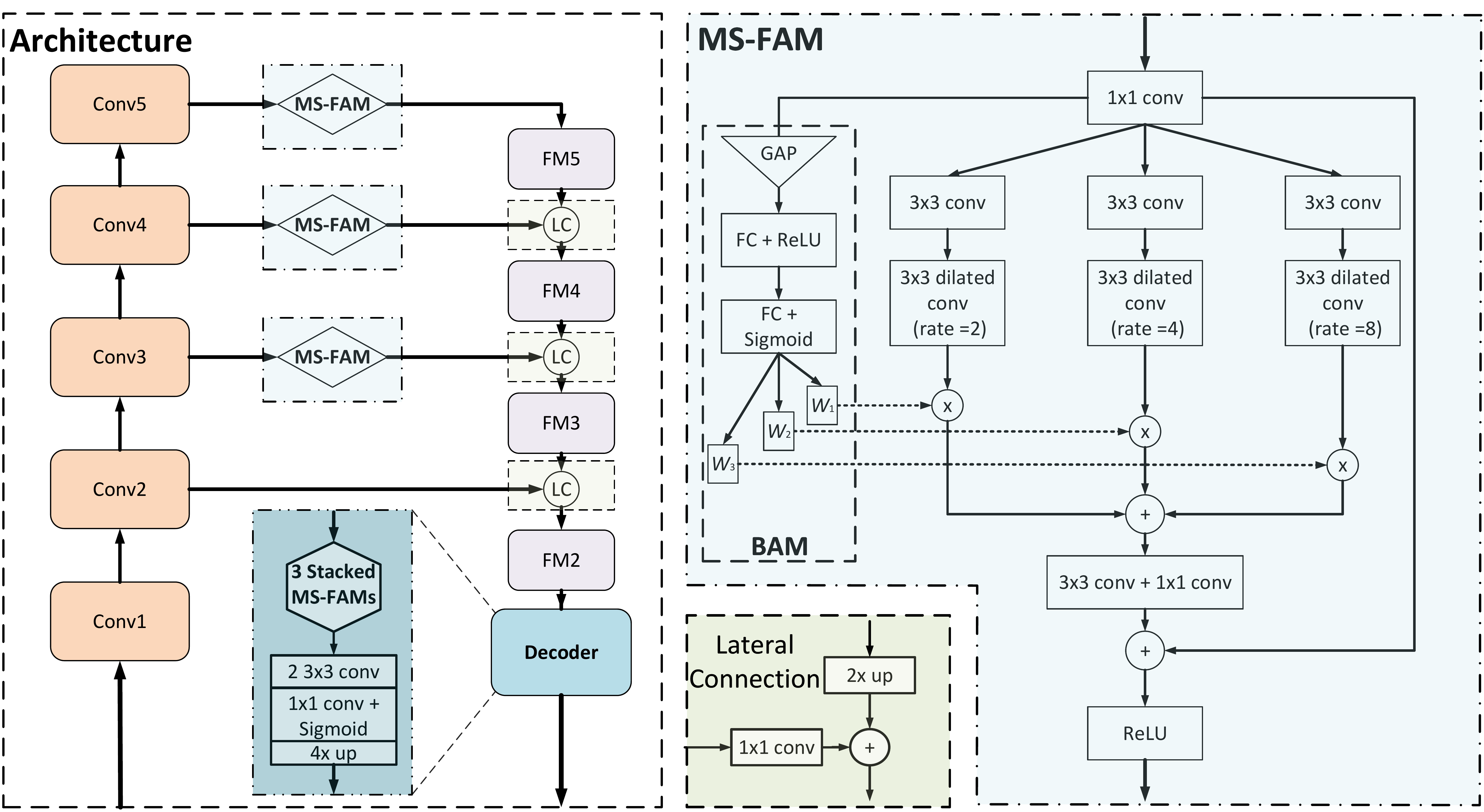}
% 	\vspace*{-6mm}
	\caption{ The overall framework of our proposed SOD model. GAP and FC are the abbreviations of global average pooling and fully connected layer, respectively. FM5 denotes the group of feature maps with the same spatial size as the output of Conv5, and so on.}
	\label{fig:images2b}
% 	\vspace{-6mm}
	% 	\vspace{-4mm}
\end{figure*} 

\subsubsection{Morphological Closing Operation and Our Simulated Version}

Morphology refers to a set of image processing operations that process images based on shapes \cite{haralick1987image}. In a morphological operation, a structuring element is applied as the comparison window to compare each pixel in the input image with its neighbors for generating the output image in a sliding window manner. There are two basic morphological operations: dilation and erosion. The former adds the foreground pixels to the boundaries of objects, while the morphological erosion removes pixels on the object boundaries. The closing operation means dilation followed by erosion operation which can be used in removing small holes inside the foreground objects. This closing operation can be directly applied to the results of existing SOD models presented in Fig.1 as a kind of post-processing method for completing them. However, it is time-consuming to perform this post-processing for every prediction as the optimized size of the structuring element is not unified in the test datasets. More importantly, the post-processing method does not always yield better results. For some salient objects with clear but close boundaries, the closing operation may merge them wrongly. Therefore, we prefer to embed this post-processing with proper modification into the training loss for creating the correct and progressive training supervisions to teach the SOD models.

The morphological dilation operation can be directly replaced by the max-pooling operation where the kernel in the max-pooling is exactly the same as the structuring element in the dilation operation. However, the erosion operation doesn't have ready-made alternatives. Erosion operation is used to shrink the object regions which are enlarged by the dilation operation. If we discard the erosion operation, some of the regions in $SM_{pgt}$ will be in the outside of $SM_{gt}$, which results in wrong training guidance. After careful consideration, we decide to use the intersect operation between the dilated network predictions and $SM_{gt}$ as the approximate alternative function of the erosion operation. There are two advantages to adopting this approximated operation: on the one hand, using $SM_{gt}$ to intersect with the dilated network predictions can always maintain the relationship of $SM_{pgt} \subseteq SM_{gt}$. It means that the generated auxiliary supervisions are correct. Using the strict closing operation to process $SM_{pred}$ may break the above relationship as it can not remove the wrongly detected pixels in the current prediction. On the other hand, our simulated morphological closing operation can always create a better result, i.e. $SM_{pgt}$, based on the shape of current prediction. This result can be used to guide the model to explore the neighboring regions of the current prediction step-by-step. As such, the spatial dependencies of salient object pixels are characterized in our PSG loss. The network will be better when its optimized results approach $SM_{pgt}$ more.

To sum up, our simulated morphological closing operation can be described by:
\begin{equation}
f(SM_{pred}) = e(d(SM_{pred})) \approx maxpool(SM_{pred}) \cap SM_{gt},
\end{equation}
where dilation operation $d(\cdot)$ is equal to the max-pooling operation $maxpool(\cdot)$ and erosion operation $e(\cdot)$ is approximated by the intersect operation with the $SM_{gt}$.

Fig. \ref{fig:images3} presents a visual example to illustrate the effectiveness of our PSG loss. In PSG loss, the SOD model will be encouraged to detect the mis-detected pixels in the neighboring regions of the current predictions with a higher priority. This is because the loss penalty weight in these pixels is (1+ $\alpha$) rather than 1 in other mis-detected yet separate pixels. As a result, the spatial dependencies of salient object pixels are characterized in our PSG loss and the SOD model can be guided to fit the training samples in such a progressive region growing way. More importantly, for a single ideal training step, the optimized results by using our PSG loss with the main loss together can be better than the results by using the main loss only. This claim is proved in the appendix of this paper.

\subsection{Architecture Overview}

An overview of our proposed SOD architecture is depicted in Fig. \ref{fig:images2b}.  Our model consists of three key components: feature pyramid network (FPN) \cite{lin2017feature}, multi-scale feature aggregation module (MS-FAM), and a decoder network.

\subsubsection{Feature Pyramid Network}  
We apply an FPN-like framework as the encoder network for feature extraction. This framework contains a bottom-up pathway, a top-down pathway, and multiple lateral connections. The bottom-up pathway (from Conv1 to Conv5) is the stage for multi-level feature extraction. The output of the last layer of each convolutional block (except Conv1) in this pathway will be used as the source of feature maps for the feature aggregation in the top-down pathway (from FM5 to FM2) via using lateral connections (LCs). Each LC aggregates feature maps of the same level (spatial size) from the bottom-up pathway and the top-down pathway. Instead of making multi-level predictions like in the original FPN, we only use the final aggregated feature maps in FM2 to feed the decoder network for saving the inference time. 

\subsubsection{Multi-scale Feature Aggregation Module}
To explicitly incorporate the multi-scale features for further improvement, we insert several multi-scale feature aggregation modules (MS-FAMs) before the LCs in our encoder network. In MS-FAM, the local context information at different scales are extracted by using parallel dilated convolutions with different dilation rates. These dilated convolutions are attached to a simple inception module \cite{yang2019dilated} which has two convolutional layers for feature dimension reduction and feature adaption. For enhancing the discrimination ability of the aggregated features, we propose to apply a branch-wise attention mechanism (BAM) to characterize the respective importance of these multi-scale local features adaptively.  It is achieved by adapting a SE block \cite{hu2018senet} from modeling the channel-wise feature dependencies to measuring the importance of each dilated branch individually. Specifically, the outputs of this attention mechanism are three separate scalars ($W_1, W_2, W_3$) in the range of 0 to 1. These branch-wise scalars are the learned weights for their corresponding dilated branches to make the sum-aggregated features more discriminative and adaptive.  The sum-aggregated features after this BAM are further processed by two additional convolutional layers for adjusting these features. Moreover, a residual connection \cite{he2016deep} is applied at the end of our MS-FAM to make it easier to optimize.

In general, our MS-FAM is used to further refine the feature maps obtained from the convolutional blocks in FPN for generating more powerful multi-scale feature maps of the same tensor size. Since the MS-FAM can maintain the tensor size of the input features, the insertion location of this module can be flexible in a fully convolutional network (FCN). In this work, three individual MS-FAMs are inserted at three connection paths between Conv3-5 and FM3-5, as depicted in Fig. \ref{fig:images2b}. To save the memory footprint, features from the Conv2 are directly fed into the FM2 group without using MS-FAMs.

\subsubsection{Decoder Network} The decoder network in our framework is used to convert the aggregated features into saliency maps. Due to the plug and play capability of our MS-FAM, this module can be incorporated in the commonly used decoder network for making it powerful. The powerful decoder network used in our paper consists of three MS-FAMs that are sequentially stacked. Besides, two $3 \times 3$ convolutional layers, one $1 \times 1$ convolutional layer, and a final bilinear up-sampling layer are used for the final saliency generation.

\section{Experiments}
\label{sec:exp}

\subsection{Experimental Setups}
\subsubsection{SOD Datasets}

Our experiments are conducted on six widely used SOD benchmark datasets: ECSSD \cite{shi2015hierarchical}, PASCAL-S \cite{li2014secrets}, DUT-O \cite{yang2013saliency}, HKU-IS \cite{li2015visual}, SOD \cite{movahedi2010design}, and DUTS \cite{wang2017learning}. The detailed information of these six datasets is presented as follows:

\begin{itemize}
\item  ECSSD \cite{shi2015hierarchical} contains 1,000 images with semantically meaningful and structurally complex natural contents. 

\item  PASCAL-S \cite{li2014secrets} contains 850 images selected from the validation dataset of PASCAL VOC 2010. 

\item  DUT-O \cite{yang2013saliency} consists of 5,168 images of relatively
complex backgrounds and high content variety.

\item  HKU-IS \cite{li2015visual} contains 4,447 images with complex scenes that typically have multiple salient objects.

\item  SOD \cite{movahedi2010design} consists of 300 challenging images from the Berkeley segmentation dataset \cite{martin2001database}. Their images have multiple salient objects with low contrast.

\item  DUTS \cite{wang2017learning} is currently the largest SOD benchmark. The images in this dataset are divided into two non-overlapping subsets: DUTS-TR and DUTS-TE. The former one contains training 10,553 images and the latter has 5,019 test images. Following the recent works \cite{liu2018picanet,wang2018detect,wu2019cascaded,liu2019simple,zhao2019egnet}, we use the DUTS-TR dataset for training our SOD models and keep the remaining datasets for test.

\end{itemize}

\subsubsection{Evaluation Criteria}

There are three widely adopted evaluation metrics, including precision-recall (PR) curves, F-measure and mean absolute error (MAE), for evaluating the performance of SOD models. 

\begin{itemize}
	\item  Precision and Recall are calculated based on the comparison between the binarized saliency map and the ground truth:
	\begin{equation}
	Precision =  \frac{TP}{TP + FP},  \  Recall =  \frac{TP}{TP + FN},  
	\end{equation}
	where TP, FP, FN denote true-positive, false-positive, and false-negative, respectively. To get the binary saliency map, all of the integers ranging from 0 to 255 is applied to the raw saliency prediction as the thresholds, each threshold will return a pair of precision and recall value to form a PR curve for visualizing the model performance.

	\item  F-measure, denoted as $F_{\beta}$, comprehensively considers both precision and recall by computing the weighted harmonic mean of them:

	\begin{equation}
	F_{\beta} = \frac{(1+ \beta^2) \times Precision \times Recall}{ \beta^2 \times Precision + Recall}
	,
	\end{equation}
	where $\beta^2$ is empirically set to 0.3 to weight more on precision. The maximum $F_{\beta}$ (maxF) values are reported as done in recent works \cite{hou2017deeply,zhao2019egnet,liu2019simple}.

	\item MAE is used to measure the average pixel-wise absolute error between the predicted saliency map $SM_{pred} \in [0,1]^{W \times H}$ and its ground truth $SM_{gt} \in \{0,1\}^{W \times H}$:
	\begin{equation}
MAE = \frac{1}{W \times H} \sum_{i=1}^{W}\sum_{j=1}^{H}|SM_{pred}(i,j) - SM_{gt}(i,j)|. 
\end{equation}

\end{itemize}

\subsubsection{Implementation Details}
\label{detail}
Our SOD model and PSG loss are implemented in Pytorch. The backbone network used in our experiments is ResNet-50. The input images to our models are all resized to $352 \times 352$ for both training and test. Our PSG loss works on the resized outputs for generating the $SM_{pgt}$. The test images are resized back to their original resolutions for evaluation. In our paper, the experimental results and figures are all based on the original test image sizes. The parameter $\alpha$ used in computing the total loss is set to 1 for equally treating $\mathcal{L}_{main}$ and $\mathcal{L}_{aux}$. The feature dimension in our MS-FAMs is fixed to 64. In the basic network, three stacked MS-FAMs are degraded by $1 \times 1$ convolutional layers with ReLU.

During training, we apply random horizontal flipping to the training images for data augmentation. The weights in the backbone network are initialized from its ImageNet pre-trained model. Our models used in the experiments are trained with Adam optimizer with an initial learning rate of $5 \times 10^{-5}$. This learning rate will be scaled down by a factor of 0.1 after half of the training epochs. The batch size is set to 20 and the total number of training epoch is 99. It is worthy to mention that the average inference time of our method is 0.024s to process an image of size $352 \times 352$ by using a GTX 1080Ti GPU. The source code of our work is publicly available\footnote{https://github.com/ysyscool/PSGLoss}.

\begin{table*}[!h]
\caption{Performance comparison on six widely used SOD datasets. $\uparrow$ and $\downarrow$ denote larger and smaller is better, respectively. In each column, the best three results are marked in \textcolor{red}{red}, \textcolor{green}{green}, and \textcolor{blue}{blue}, respectively.}
\label{tab:pc}	
\begin{adjustbox}{width=1\textwidth}
	\begin{tabular}{l|c|c|c|c|c|c|c|c|c|c|c|c|c|c|c}
		\hline
		\multirow{2}{*}{Model} & \multirow{2}{*}{Backbone} & \multicolumn{2}{c|}{ECSSD \cite{shi2015hierarchical}} & \multicolumn{2}{c|}{PASCAL-S \cite{li2014secrets}} & \multicolumn{2}{c|}{DUT-O \cite{yang2013saliency}} & \multicolumn{2}{c|}{HKU-IS \cite{li2015visual}} & \multicolumn{2}{c|}{SOD \cite{movahedi2010design}} & \multicolumn{2}{c|}{DUTS-TE \cite{wang2017learning}} & \multicolumn{2}{c}{Average}\\ \cline{3-16} 
		&                           & maxF$\uparrow$         & MAE$\downarrow$         & maxF$\uparrow$         & MAE$\downarrow$            & maxF$\uparrow$         & MAE$\downarrow$          & maxF$\uparrow$         & MAE$\downarrow$          & maxF$\uparrow$        & MAE$\downarrow$         & maxF$\uparrow$         & MAE$\downarrow$   & maxF$\uparrow$         & MAE$\downarrow$        \\ \hline
		\multicolumn{16}{c}{VGG-based}  \\ \hline 
		
DCL          & VGG-16    & 0.898 & 0.078 & 0.820         & 0.113        & 0.759 & 0.086 & 0.908  & 0.055 & 0.823 & 0.193 & 0.782  & 0.148 & 0.832   & 0.112 \\
UCF          & VGG-16    & 0.902 & 0.070  & 0.816         & 0.115        & 0.730 & 0.120 & 0.888  & 0.061 & 0.800 & 0.164 & 0.772  & 0.111 & 0.818    & 0.107 \\
Amulet       & VGG-16    & 0.914 & 0.060  & 0.830        & 0.100          & 0.743 & 0.097 & 0.899  & 0.050  & 0.805 & 0.141 & 0.778  & 0.084 & 0.828   & 0.089 \\
NLDF         & VGG-16    & 0.903 & 0.065 & 0.822         & 0.098        & 0.753 & 0.079 & 0.902  & 0.048 & 0.837 & 0.123 & 0.816  & 0.065 & 0.839   & 0.080  \\
PAGRN        & VGG-19    & 0.924 & 0.064 & 0.849         & 0.089        & 0.771 & 0.071 & 0.919  & 0.047 & 0.838 & 0.145 & 0.854  & 0.055 & 0.859   & 0.079 \\

DSS          & VGG-16    & 0.919 & 0.055 & 0.833         & 0.093        & 0.781  & 0.063 & 0.916   & 0.040  & 0.843 & 0.121 & 0.825  & 0.056 & 0.853   & 0.071 \\
\hline 
\multicolumn{16}{c}{ResNet-based}  \\ \hline 

SRM          & ResNet-50 & 0.916 & 0.056 & 0.840         & 0.084        & 0.769 & 0.069 & 0.906  & 0.046 & 0.843  & 0.126 & 0.826  & 0.058 & 0.850  & 0.073 \\
PAGENet      & ResNet-50 & 0.928 & 0.046 & 0.850          & 0.076        & 0.791 & 0.062 & 0.920   & 0.036 & 0.839 & 0.110  & 0.838  & 0.051 & 0.861   & 0.064 \\
PiCANet      & ResNet-50 & 0.932 & 0.048 & 0.864         & 0.075        & 0.820  & 0.064 & 0.920   & 0.044 & 0.861 & \textcolor{blue}{0.103} & 0.863  & 0.05  & 0.877   & 0.064 \\
DGRL         & ResNet-50 & 0.921 & 0.043 & 0.844         & 0.075        & 0.774 & 0.062 & 0.910   & 0.036 & 0.843 & \textcolor{blue}{0.103} & 0.828  & 0.049 & 0.853   & 0.061 \\
BASNet       & ResNet-34 & 0.939 & 0.040  & 0.858         & 0.076        & 0.811 & 0.057 & 0.930   & 0.033 & 0.851 & 0.112 & 0.860   & 0.047 & 0.875   & 0.061 \\
CPD          & ResNet-50 & 0.936 & 0.040  & 0.861         & \textcolor{blue}{0.071}        & 0.796 & 0.056 & 0.928  & 0.033 & 0.859 & 0.110  & 0.865  & 0.043 & 0.874   & 0.059 \\
PoolNet      & ResNet-50 & 0.939 & 0.043 & 0.867         & 0.074        & \textcolor{blue}{0.826} & \textcolor{green}{0.054} & 0.932  & 0.033 & 0.866 & 0.109 & 0.883  & \textcolor{blue}{0.039} & 0.886   & 0.059 \\
EGNet        & ResNet-50 & \textcolor{blue}{0.942} & \textcolor{blue}{0.041} & \textcolor{blue}{0.872}         & 0.074        & \textcolor{red}{0.843} & 0.055 & \textcolor{green}{0.936}  & \textcolor{blue}{0.032} & \textcolor{green}{0.873} & 0.106 & \textcolor{red}{0.894}  & \textcolor{blue}{0.039} & \textcolor{red}{0.893}   & \textcolor{blue}{0.058} \\\hline
Ours(BCE)    & ResNet-50 & \textcolor{green}{0.944} & \textcolor{green}{0.039} & \textcolor{green}{0.875}         & \textcolor{green}{0.065}        & 0.820  & \textcolor{green}{0.054} & \textcolor{green}{0.936}  & \textcolor{green}{0.030}  & \textcolor{red}{0.879} & \textcolor{green}{0.102} & \textcolor{blue}{0.886}  & \textcolor{green}{0.037} & \textcolor{blue}{0.890}    & \textcolor{green}{0.055} \\
Ours(Hybrid) & ResNet-50 & \textcolor{red}{0.946} & \textcolor{red}{0.035} & \textcolor{red}{0.879}         & \textcolor{red}{0.061}        & \textcolor{green}{0.828} & \textcolor{red}{0.053} & \textcolor{red}{0.938}  & \textcolor{red}{0.027} & \textcolor{blue}{0.872} & \textcolor{red}{0.096} & \textcolor{green}{0.890}   & \textcolor{red}{0.036} & \textcolor{green}{0.892}   & \textcolor{red}{0.051}   

\\ \hline
	\end{tabular}
\end{adjustbox}
\end{table*}

\begin{figure*}[!h]
	\centering
% 	\vspace*{-6mm}
	\includegraphics[page=4,trim = 5mm 5mm 5mm 5mm, clip, width=1.0\linewidth]{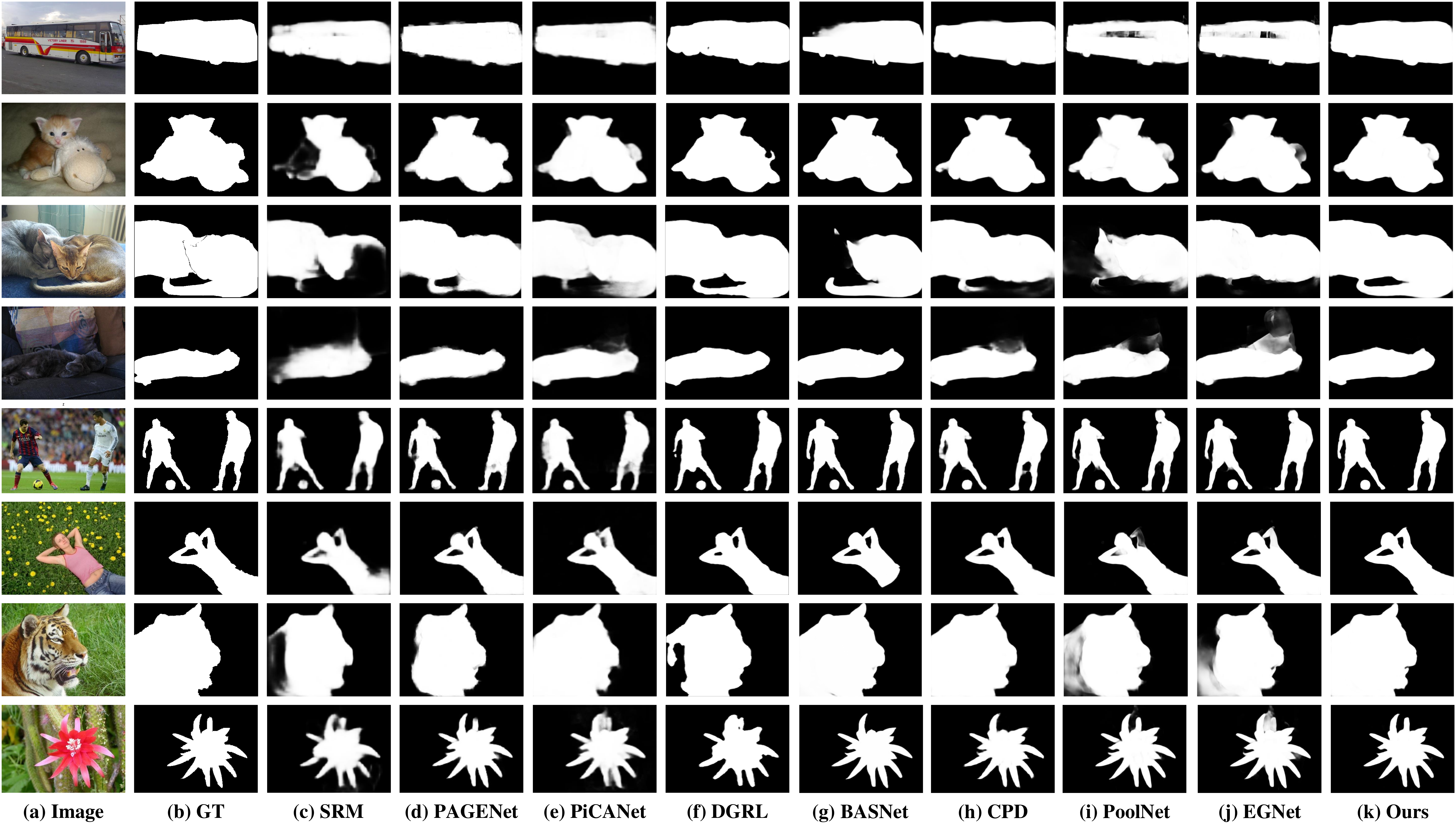}
% 	\vspace*{-6mm}
	\caption{ Visual comparisons of our method and other ResNet-based models on some representative examples (best viewed digitally with zoom).}
	\label{fig:images4}
% 	\vspace{-6mm}
	% 	\vspace{-4mm}
\end{figure*}

\begin{figure*}[!h]
	\centering
% 	\vspace*{-6mm}
		\includegraphics[trim = 25mm 1mm 20mm 1mm, clip, width=0.24\linewidth]{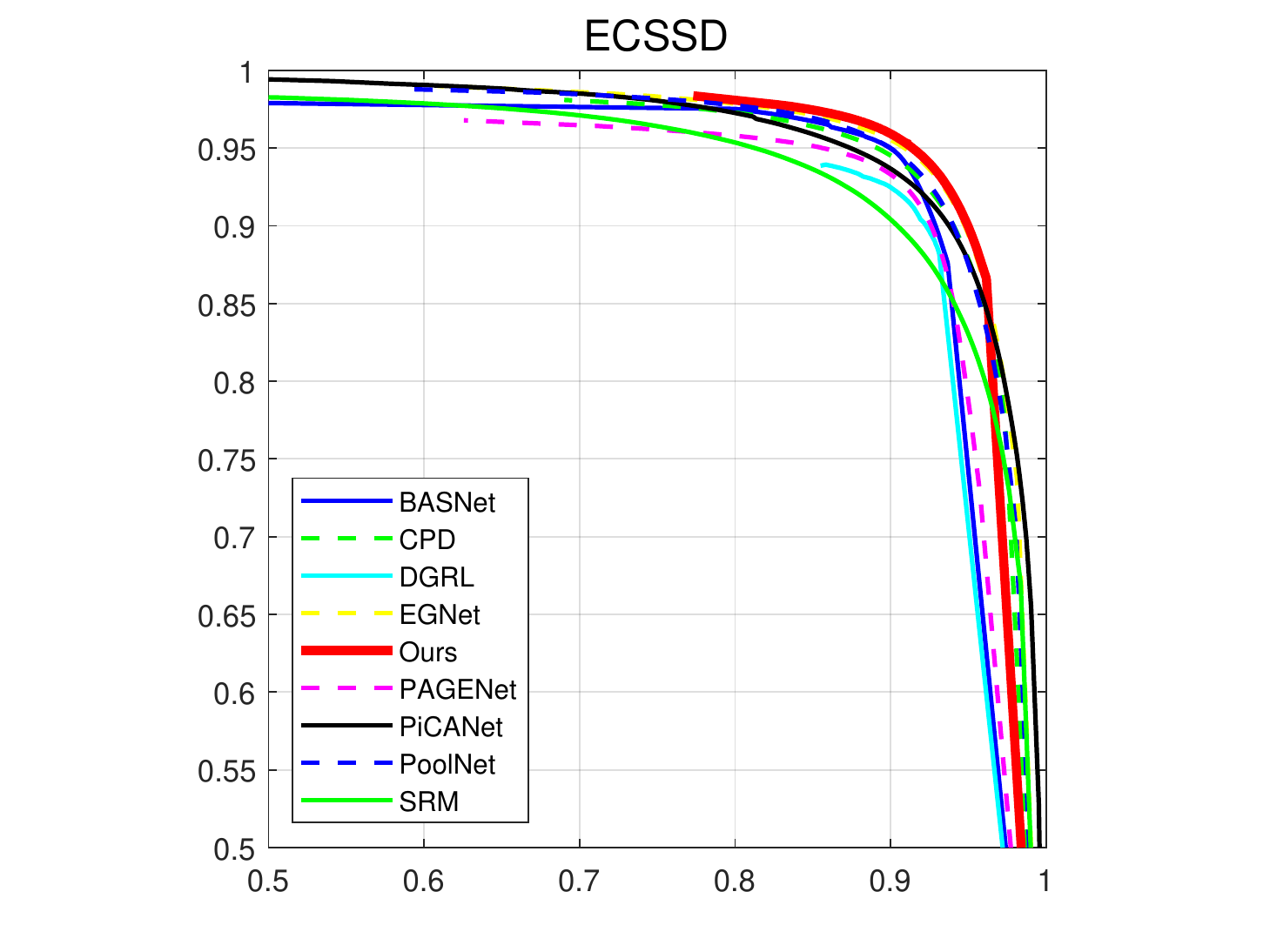}\hfill
		\includegraphics[trim = 25mm 1mm 20mm 1mm, clip, width=0.24\linewidth]{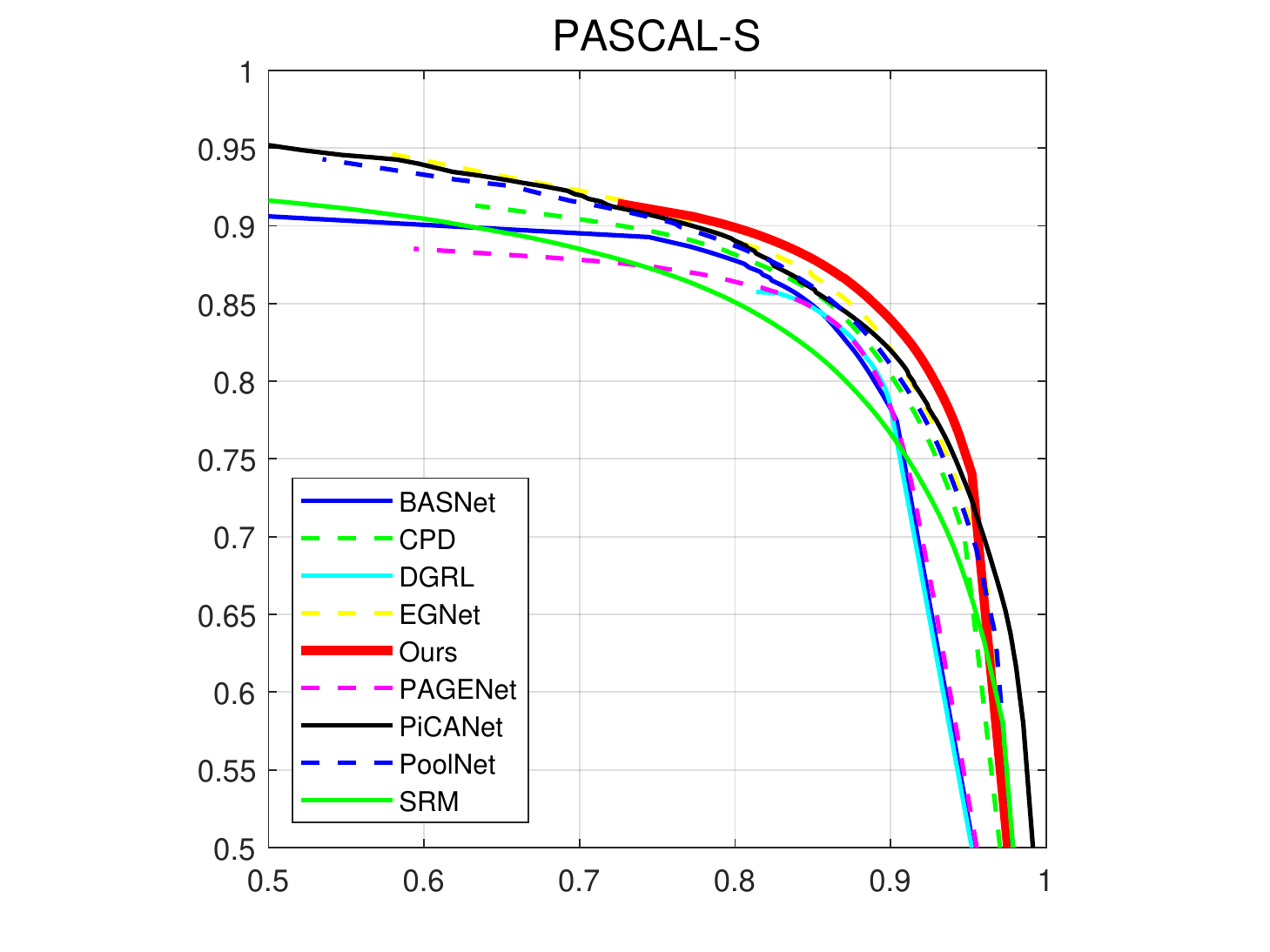} \hfill
		\includegraphics[trim = 25mm 1mm 20mm 1mm, clip, width=0.24\linewidth]{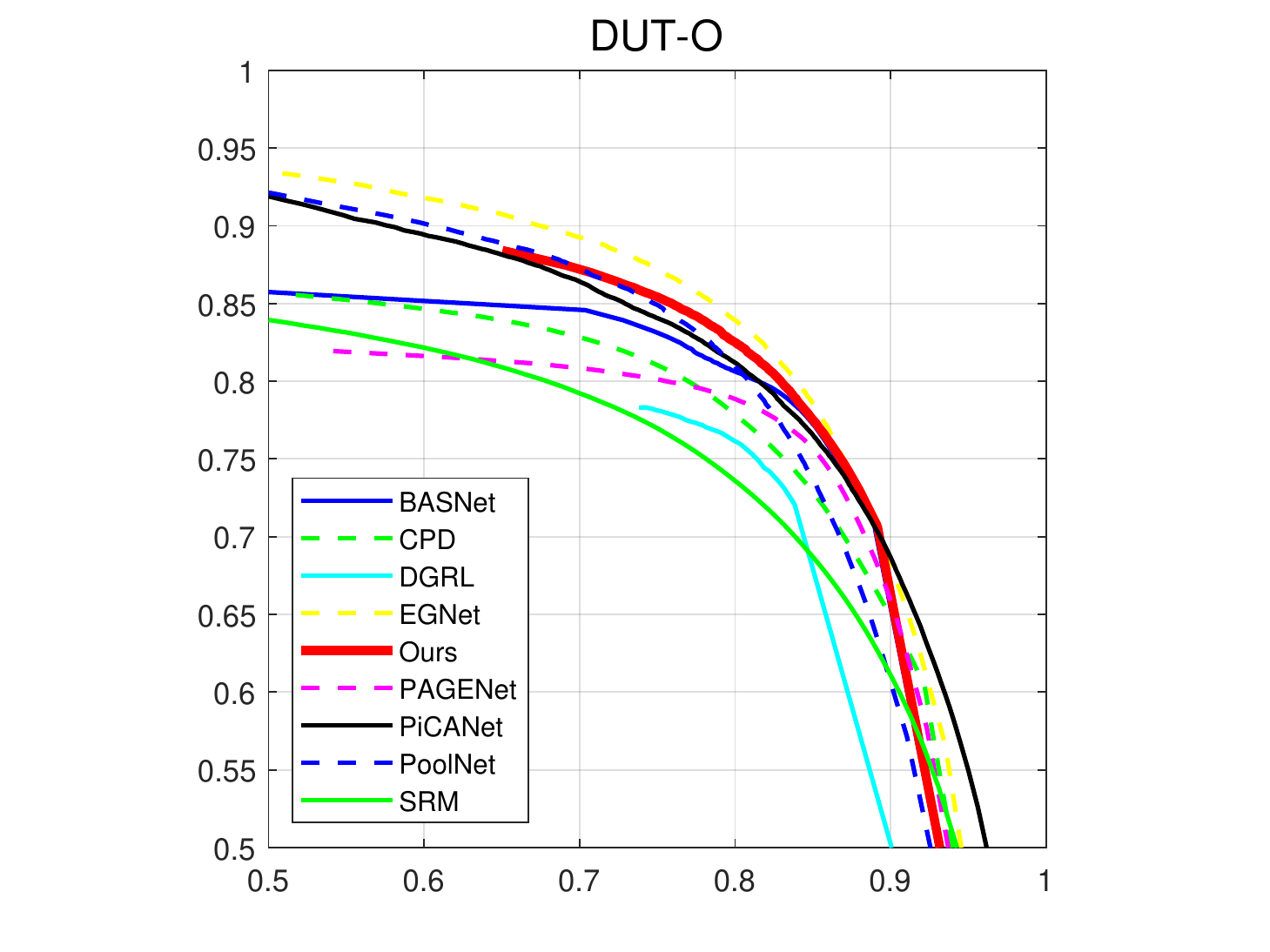} \hfill
		\includegraphics[trim = 25mm 1mm 20mm 1mm, clip, width=0.24\linewidth]{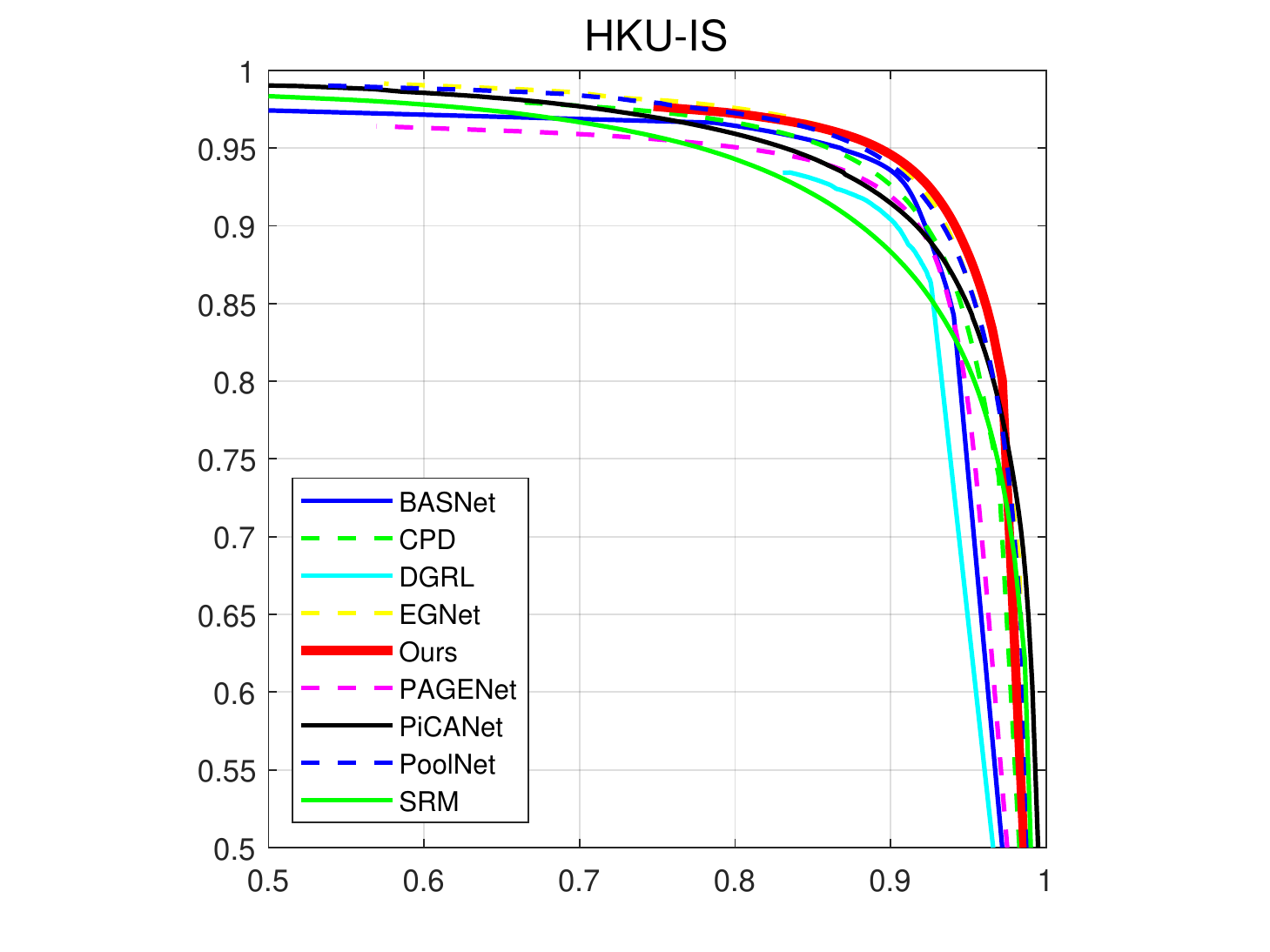}

\caption{Performance comparison with PR-curve on four SOD benchmarks. PR-curve: precision (vertical axis) recall (horizontal axis) curve. Our model obtains promising performance on three of them. Best viewed in color. }
	\label{fig:imageszz}
	
% 	\vspace{-6mm}
	% 	\vspace{-4mm}
\end{figure*}

\begin{table*}[ht]
\caption{Model ablation analysis of our method with maxF (higher is better) and MAE (lower is better) on six SOD benchmarks. In each column, the best two results are marked in \textcolor{red}{red} and \textcolor{green}{green}, respectively. + in Encoder or Decoder denote MS-FAMs are incorporated. }
\label{tab:ma}

\begin{adjustbox}{width=1\textwidth}
	\begin{tabular}{c|c|c|c|c|c|c|c|c|c|c|c|c|c|c|c|c}% |cc|cc
	\hline
\multicolumn{2}{c|}{Model Variants}                                                                                                                                                                       & \multicolumn{1}{c|}{\multirow{2}{*}{PSG Loss}} & \multicolumn{2}{c|}{ECSSD \cite{shi2015hierarchical}} & \multicolumn{2}{c|}{PASCAL-S \cite{li2014secrets}} & \multicolumn{2}{c|}{DUT-O \cite{yang2013saliency}} & \multicolumn{2}{c|}{HKU-IS \cite{li2015visual}} & \multicolumn{2}{c|}{SOD \cite{movahedi2010design}} & \multicolumn{2}{c|}{DUTS-TE \cite{wang2017learning}}  & \multicolumn{2}{c}{Average}            \\ \cline{1-2} \cline{4-17}  
Encoder           & Decoder          &                           & maxF$\uparrow$         & MAE$\downarrow$         & maxF$\uparrow$         & MAE$\downarrow$            & maxF$\uparrow$         & MAE$\downarrow$          & maxF$\uparrow$         & MAE$\downarrow$          & maxF$\uparrow$        & MAE$\downarrow$         & maxF$\uparrow$         & MAE$\downarrow$   & maxF$\uparrow$         & MAE$\downarrow$         \\ \hline

	\multicolumn{17}{c}{Effectiveness of MS-FAM}  \\ \hline   

                &                    &         & 0.935 & 0.043 & 0.867         & 0.067        & 0.811 & 0.058 & 0.931  & 0.031 & 0.859 & 0.111 & 0.875  & 0.040  & 0.880    & 0.058  \\
                & +                 &         & 0.938 & 0.042 & 0.872         & 0.066        & 0.816 & 0.056 & 0.931  & 0.031 & \textcolor{green}{0.874} & 0.109 & 0.878  & 0.039 & 0.885   & 0.057  \\
+          &                    &         & 0.941 & 0.038 & 0.870          & 0.066        & 0.824 & 0.054 & 0.934  & 0.029 & 0.867 & \textcolor{green}{0.101} & 0.881  & 0.038 & 0.886   & 0.054\\
+              & +               &         & 0.941 & 0.037 & 0.874         & \textcolor{green}{0.063}        & 0.823 & 0.055 & \textcolor{green}{0.937}  & \textcolor{green}{0.028} & \textcolor{red}{0.879} & 0.102 & 0.883  & 0.037 & \textcolor{green}{0.890}    & 0.054 \\ \hline

		\multicolumn{17}{c}{Effectiveness of PSG Loss}  \\ \hline                                                                                                                                                                       
                &                    & +        & 0.937 & 0.043 & 0.866         & 0.068        & 0.816 & 0.055 & 0.934  & 0.030  & 0.871 & 0.108 & 0.879  & 0.039 & 0.884   & 0.057 \\
                & +                  & +        & 0.940  & 0.040  & \textcolor{green}{0.877}         & 0.065        & \textcolor{green}{0.826} & \textcolor{red}{0.053} & 0.934  & 0.030  & 0.869 & 0.107 & 0.884  & 0.038 & 0.888   & 0.056 \\
+               &                    & +       & \textcolor{green}{0.944} & \textcolor{green}{0.036} & 0.876         & \textcolor{green}{0.063}        & 0.821 & \textcolor{red}{0.053} & 0.935  & 0.029 & 0.871 & \textcolor{green}{0.101} & \textcolor{green}{0.887}  & \textcolor{red}{0.036} & 0.889   & \textcolor{green}{0.053} \\
+               & +                 & +       & \textcolor{red}{0.946} & \textcolor{red}{0.035} & \textcolor{red}{0.879}         & \textcolor{red}{0.061}        & \textcolor{red}{0.828} & \textcolor{red}{0.053} & \textcolor{red}{0.938}  & \textcolor{red}{0.027} & 0.872 & \textcolor{red}{0.096} & \textcolor{red}{0.890}   & \textcolor{red}{0.036} & \textcolor{red}{0.892}   & \textcolor{red}{0.051}  \\ \hline
	\end{tabular}
\end{adjustbox}
\end{table*}

\subsection{Performance Comparison}
We compare our method with 14 recent SOD models: DCL \cite{li2016deep}, UCF \cite{zhang2017learning}, Amulet \cite{zhang2017amulet}, PAGRN \cite{zhang2018progressive}, DSS \cite{hou2017deeply}, NLDF \cite{luo2017non}, PiCANet \cite{liu2018picanet}, SRM \cite{wang2017stagewise}, PAGENet \cite{wang2019salient},  DGRL \cite{wang2018detect}, BASNet \cite{qin2019basnet}, CPD \cite{wu2019cascaded}, PoolNet \cite{liu2019simple} and EGNet \cite{zhao2019egnet}. Among them, the former six models are VGG-based while the latter eight models are ResNet-based. For fair comparisons, all the saliency maps of the above methods are released by the authors or generated by using the available source codes. 

In Table \ref{tab:pc}, we present the quantitative performance comparison results on the six SOD datasets, as well as the average results. For maxF and MAE scores on each dataset, the best three models are highlighted in \textcolor{red}{red}, \textcolor{green}{green}, and \textcolor{blue}{blue}, respectively. From this table, we can observe that our method achieves the state-of-the-art performance on almost all benchmark datasets, in terms of MAE. As for maxF, our method achieves the best two performances on most benchmarks. EGNet \cite{zhao2019egnet} has the optimal average maxF due to its outstanding F-measure performance on DUT-O and DUTS-TE datasets. It should be credited to the additional edge mask labels used in this method. The results of our model with BCE loss only are also presented in this table. Our method can still outperform the existing works on ECSSD, PASCAL-S, HKU-IS, and SOD datasets, in terms of both maxF and MAE. By utilizing the extra Dice loss, our model can further improve the performance on almost all datasets, except the SOD dataset. We guess that the SOD dataset only has 300 images for testing which may have a different label distribution from the training images in the DUTS-TR dataset. Fig.\ref{fig:images4} shows a visual comparison of the results of our method against other ResNet-based models. As we can see that incomplete prediction is a common and unaddressed problem and the detected salient objects by using our method are more complete than other competitors.

Besides, the precision-recall curves of all ResNet-based methods on the first four datasets are provided in Fig. \ref{fig:imageszz}. We can see that our method outperforms its counterparts on ECSSD, PASCAL-S, and HKU-IS datasets. While for the DUT-O dataset, our model is the best except for EGNet \cite{zhao2019egnet}. The additional saliency edge labels used in EGNet contribute mainly to its advantage. It should be emphasized that the proposed PSG loss has helped our architecture greatly on these datasets. If we plot the PR curves of our model without trained with PSG loss, worse curves will be obtained.

\subsection{Ablation Study}

In this subsection, we conduct a series of ablation experiments to analyze the contribution of two key components, including the multi-scale feature aggregation module (MS-FAM) and progressive self-guided (PSG) loss, in our method. The quantitative results of our ablation study on the six SOD benchmarks are summarized in Table \ref{tab:ma}. Our baseline model, as shown in the first row of this table, consists of an FPN-like encoder and a fully convolutional decoder network without MS-FAMs.  Our MS-FAM is a flexible convolutional module for extracting and fusing the multi-scale features, which can be applied in both encoder and decoder networks. As a result, we have four model variants as the subjects in this ablation study.

\begin{table*}[!ht]
\caption{Performance comparison of the training losses and the losses combined with our PSG loss.  For each loss coupled with our PSG loss, the improved or degraded results are marked in \textcolor{red}{red} or \textcolor{green}{green}, respectively.}
\label{tab:maloss}	
\begin{adjustbox}{width=1\textwidth}

\begin{tabular}{l|c|c|c|c|c|c|c|c|c|c|c|c|c|c|c}
\hline
\multirow{2}{*}{Loss}     & \multirow{2}{*}{PSG Loss} & \multicolumn{2}{c|}{ECSSD \cite{shi2015hierarchical}} & \multicolumn{2}{c|}{PASCAL-S \cite{li2014secrets}} & \multicolumn{2}{c|}{DUT-O \cite{yang2013saliency}} & \multicolumn{2}{c|}{HKU-IS \cite{li2015visual}} & \multicolumn{2}{c|}{SOD \cite{movahedi2010design}} & \multicolumn{2}{c|}{DUTS-TE \cite{wang2017learning}} & \multicolumn{2}{c}{Average}  \\ \cline{3-16} 
                          &                           & maxF$\uparrow$         & MAE$\downarrow$            & maxF$\uparrow$         & MAE$\downarrow$          & maxF$\uparrow$         & MAE$\downarrow$          & maxF$\uparrow$        & MAE$\downarrow$         & maxF$\uparrow$         & MAE$\downarrow$   & maxF$\uparrow$         & MAE$\downarrow$    & maxF$\uparrow$         & MAE$\downarrow$        \\ \hline
\multirow{2}{*}{$\ell1$-norm}       &                          & 0.940  & 0.037 & 0.870          & 0.066        & 0.820  & 0.054 & 0.933  & 0.029 & 0.872 & 0.103 & 0.876  & 0.039 & 0.885   & 0.055      \\  
                          & +                         & \textcolor{red}{0.944} & \textcolor{red}{0.036} & \textcolor{red}{0.876}         & \textcolor{red}{0.062}        &\textcolor{red}{0.823} & \textcolor{red}{0.052} & 0.933  & \textcolor{red}{0.028} & \textcolor{green}{0.865} & \textcolor{red}{0.100}   & \textcolor{red}{0.882}  & \textcolor{red}{0.036} & \textcolor{red}{0.887}   & \textcolor{red}{0.052}       \\ \hline
\multirow{2}{*}{$\ell2$-norm}       &                          & 0.935 & 0.045 & 0.869         & 0.068        & 0.816 & 0.057 & 0.930   & 0.035 & 0.868 & 0.110  & 0.873  & 0.042 & 0.882   & 0.060        \\ 
                          & +                        & \textcolor{red}{0.939} & \textcolor{red}{0.044} & 
                          \textcolor{red}{0.872}         & 
                          \textcolor{red}{0.066}        & 0.816 & \textcolor{red}{0.055} & \textcolor{red}{0.932}  & \textcolor{red}{0.034} & 0.868 & \textcolor{green}{0.114} & \textcolor{red}{0.874}  & \textcolor{red}{0.041} & \textcolor{red}{0.884}   & \textcolor{red}{0.059}      \\ \hline
\multirow{2}{*}{KLD}      &                          & 0.943 & 0.040  & 0.876         & 0.066        & 0.826 & 0.055 & 0.935  & 0.031 & 0.878 & 0.100   & 0.884  & 0.039 & 0.890    & 0.055      \\ 
                          & +                         & \textcolor{red}{0.944} & \textcolor{green}{0.041} & \textcolor{red}{0.877}         & \textcolor{red}{0.065}        & \textcolor{red}{0.829} & \textcolor{red}{0.053} & \textcolor{red}{0.937}  & \textcolor{red}{0.030}  & \textcolor{red}{0.881} & \textcolor{green}{0.103} & \textcolor{red}{0.887}  & \textcolor{red}{0.038} & \textcolor{red}{0.893}   & 0.055       \\ \hline
\multirow{2}{*}{Dice}     &                          & 0.936 & 0.039 & 0.863         & 0.065        & 0.819 & 0.053 & 0.926  & 0.029 & 0.862 & 0.101 & 0.874  & 0.038 & 0.880    & 0.054    \\ 
                          & +                         & \textcolor{red}{0.938} & \textcolor{red}{0.037} & \textcolor{green}{0.862}         & 0.065        & 0.819 & \textcolor{red}{0.052} & \textcolor{red}{0.929}  & \textcolor{red}{0.027} & 0.862 & \textcolor{red}{0.099} & \textcolor{red}{0.877}  & \textcolor{red}{0.037} & \textcolor{red}{0.881}   & \textcolor{red}{0.053}    \\ \hline
\multirow{2}{*}{BCE}      &                          &  0.941 & 0.040  & 0.870          & 0.066        & 0.820  & 0.054 & 0.937  & 0.030  & 0.867 & 0.105 & 0.885  & 0.037 & 0.887   & 0.055      \\ 
                          & +                          & \textcolor{red}{0.944} & \textcolor{red}{0.039} & \textcolor{red}{0.875}         & \textcolor{red}{0.065}        & 0.820  & 0.054 & \textcolor{green}{0.936}  & 0.030  & \textcolor{red}{0.879} & \textcolor{red}{0.102} & \textcolor{red}{0.886}  & 0.037 & \textcolor{red}{0.890}    & 0.055       \\ \hline
\multirow{2}{*}{BCE + Dice} &                          & 0.941 & 0.037 & 0.874         & 0.063        & 0.823 & 0.055 & 0.937  & 0.028 & 0.879 & 0.102 & 0.883  & 0.037 & 0.890    & 0.054      \\  
                          & +                      & \textcolor{red}{0.946} & \textcolor{red}{0.035} & \textcolor{red}{0.879}         & \textcolor{red}{0.061}        & \textcolor{red}{0.828} & \textcolor{red}{0.053} & \textcolor{red}{0.938}  & \textcolor{red}{0.027} & \textcolor{green}{0.872} & \textcolor{red}{0.096} & \textcolor{red}{0.890}   & \textcolor{red}{0.036} & \textcolor{red}{0.892}   & \textcolor{red}{0.051}       \\ \hline
\end{tabular}
\end{adjustbox}
\end{table*}

\begin{table*}[!h]
\caption{Performance comparison of the original models and the models retrained with our PSG loss. + denotes the retrained one. For each retrained model, the improved or degraded results are marked in \textcolor{red}{red} or \textcolor{green}{green}, respectively.}
\label{tab:ma2}	
\begin{adjustbox}{width=1\textwidth}
	\begin{tabular}{l|c|c|c|c|c|c|c|c|c|c|c|c|c|c}
		\hline
		\multirow{2}{*}{Model} & \multicolumn{2}{c|}{ECSSD \cite{shi2015hierarchical}} & \multicolumn{2}{c|}{PASCAL-S \cite{li2014secrets}} & \multicolumn{2}{c|}{DUT-O \cite{yang2013saliency}} & \multicolumn{2}{c|}{HKU-IS \cite{li2015visual}} & \multicolumn{2}{c|}{SOD \cite{movahedi2010design}} & \multicolumn{2}{c|}{DUTS-TE \cite{wang2017learning}}  & \multicolumn{2}{c}{Average}   \\ \cline{2-15} 
	      & maxF$\uparrow$         & MAE$\downarrow$            & maxF$\uparrow$         & MAE$\downarrow$          & maxF$\uparrow$         & MAE$\downarrow$          & maxF$\uparrow$        & MAE$\downarrow$         & maxF$\uparrow$         & MAE$\downarrow$   & maxF$\uparrow$         & MAE$\downarrow$    & maxF$\uparrow$         & MAE$\downarrow$        \\ \hline

CPD                    & 0.936 & 0.040  & 0.861         & 0.071        & 0.796 & 0.056 & 0.928  & 0.033 & 0.859 & 0.110  & 0.865  & 0.043 & 0.874   & 0.059 \\
CPD+                   & \textcolor{red}{0.939} & 0.040  & \textcolor{red}{0.864}         & \textcolor{red}{0.067}        & \textcolor{red}{0.801} & \textcolor{red}{0.055} & \textcolor{red}{0.929}  & \textcolor{red}{0.032} & \textcolor{red}{0.859} & \textcolor{red}{0.106} & \textcolor{red}{0.868}  & \textcolor{red}{0.042} & \textcolor{red}{0.877}   & \textcolor{red}{0.057}  \\ \hline
PoolNet                & 0.939 & 0.043 & 0.867         & 0.074        & 0.826 & 0.054 & 0.932  & 0.033 & 0.866 & 0.109 & 0.883  & 0.039 & 0.886   & 0.059 \\
PoolNet+               & \textcolor{red}{0.940}  & \textcolor{red}{0.041} & \textcolor{red}{0.870}          & \textcolor{red}{0.071}        & \textcolor{red}{0.830}  & 0.054 & \textcolor{red}{0.935}  & \textcolor{red}{0.031} & \textcolor{red}{0.875} & \textcolor{red}{0.105} & \textcolor{red}{0.887}  & 0.039 & \textcolor{red}{0.890}    & \textcolor{red}{0.057}  \\\hline
EGNet                 & 0.942 & 0.041 & 0.872         & 0.074        & 0.843 & 0.055 & 0.936  & 0.032 & 0.873 & 0.106 & 0.894  & 0.039 & 0.893   & 0.058 \\
EGNet+                 & \textcolor{red}{0.945} & \textcolor{red}{0.039} & \textcolor{red}{0.873}         & \textcolor{red}{0.071}        & \textcolor{green}{0.841} & \textcolor{red}{0.052} & \textcolor{red}{0.938}  & \textcolor{red}{0.031} & \textcolor{red}{0.878} & \textcolor{green}{0.108} & \textcolor{red}{0.895}  & \textcolor{red}{0.037} & \textcolor{red}{0.895}   & \textcolor{red}{0.056}\\ \hline

	\end{tabular}
\end{adjustbox}
\end{table*}

\subsubsection{Effectiveness of MS-FAM}
As shown in Table \ref{tab:ma}, the models equipped with MS-FAMs can easily outperform the baseline model in the most of test datasets, regardless of the application location of MS-FAMs. The average maxF and MAE scores can be improved at least 0.6\% and 1.7\%, respectively, by using MS-FAMs in our basic SOD model. These results verify the conclusion in many SOD papers \cite{hou2017deeply,li2015visual,zhang2017amulet} that the SOD performance indeed boosted by incorporating the multi-scale features. 

Moreover, the model with MS-FAMs used in the encoder network can achieve better results in the majority of test benchmarks than the one with MS-FAMs used in the decoder. It means that multi-scale features should be better incorporated into the encoder network before the saliency inference stage. The model in the fourth row of Table \ref{tab:ma} is the final architecture of our SOD model. It can be used to illustrate that making a powerful decoder network can also help in improving the SOD performance.

The branch-wise attention mechanism (BAM) plays an important role in our MS-FAM. If we discard it by only applying the parallel dilated convolutions with sum-aggregation in MS-FAM, the performance of our last model variant will lose 0.002 scores on average results, in terms of both maxF and MAE.

\subsubsection{Effectiveness of PSG Loss}
PSG loss can be applied to the above-mentioned model variants for guiding their training. From Table \ref{tab:ma}, we can observe that the model trained with PSG loss can surpass its normally trained version in most benchmarks, convincingly demonstrating the effectiveness of this loss. We note that some of the model variants trained with PSG loss will obtain a better MAE yet worse maxF result on some of the datasets. We think that it may be related to the different characteristics of these two evaluation metrics. Lower MAE means a lower pixel-wise difference between the model prediction and the corresponding ground truth. While higher maxF represents a better binary classification performance.

The kernel size of max-pooling used in our PSG loss is one of the most important hyper-parameters for designing it. The current setting is using a $3 \times 3$ max-pooling layer for the morphological dilation.  We find that the performance of using $5 \times 5$ or even larger size of max-pooling is not good as using this one, but still better than not using PSG loss. Obviously, using a larger size of max-pooling will result in a larger shape change between the $SM_{pred}$ and $SM_{pgt}$. One possible explanation is that using a larger size of max-pooling will make the $SM_{pgt}$ more quickly similar to $SM_{gt}$ in the early training epochs. It will make our PSG loss share the same utility as the main loss and lose its capability in modeling spatial dependencies.

\subsubsection{PSG Loss with other training losses}
Besides the hybrid loss (BCE + Dice) used in the existing experiments, our PSG can work with other training losses as well. We choose the fourth model variant (Model4), i.e. the model with MS-FAMs in both encoder and decoder, in our ablation study as the experiment subject in this part. Five commonly used segmentation training losses, such as $\ell1$-norm, $\ell2$-norm, KLD (Kullback–Leibler divergence), Dice, and BCE loss, are individually used to validate the generalization ability of PSG loss. The results are presented in Table \ref{tab:maloss}. For each loss coupled with our PSG loss, the improved or degraded results are marked in \textcolor{red}{red} or \textcolor{green}{green}, respectively. We can easily find that the performance of Model4 trained with a specific training loss can be boosted by incorporating the PSG loss. Especially, our model can achieve the same average maxF scores and better MAE performance than EGNet by using the combination of KLD loss and our PSG loss. But its individual performance on many benchmarks is not as remarkable as our final chosen model (trained with the hybrid loss and PSG loss).

\subsubsection{Post-processing vs. PSG Loss} 

\begin{figure}[ht]
	\centering
% 	\vspace*{-6mm}
	\includegraphics[page=6,trim = 5mm 5mm 5mm 5mm, clip, width=1.0\linewidth]{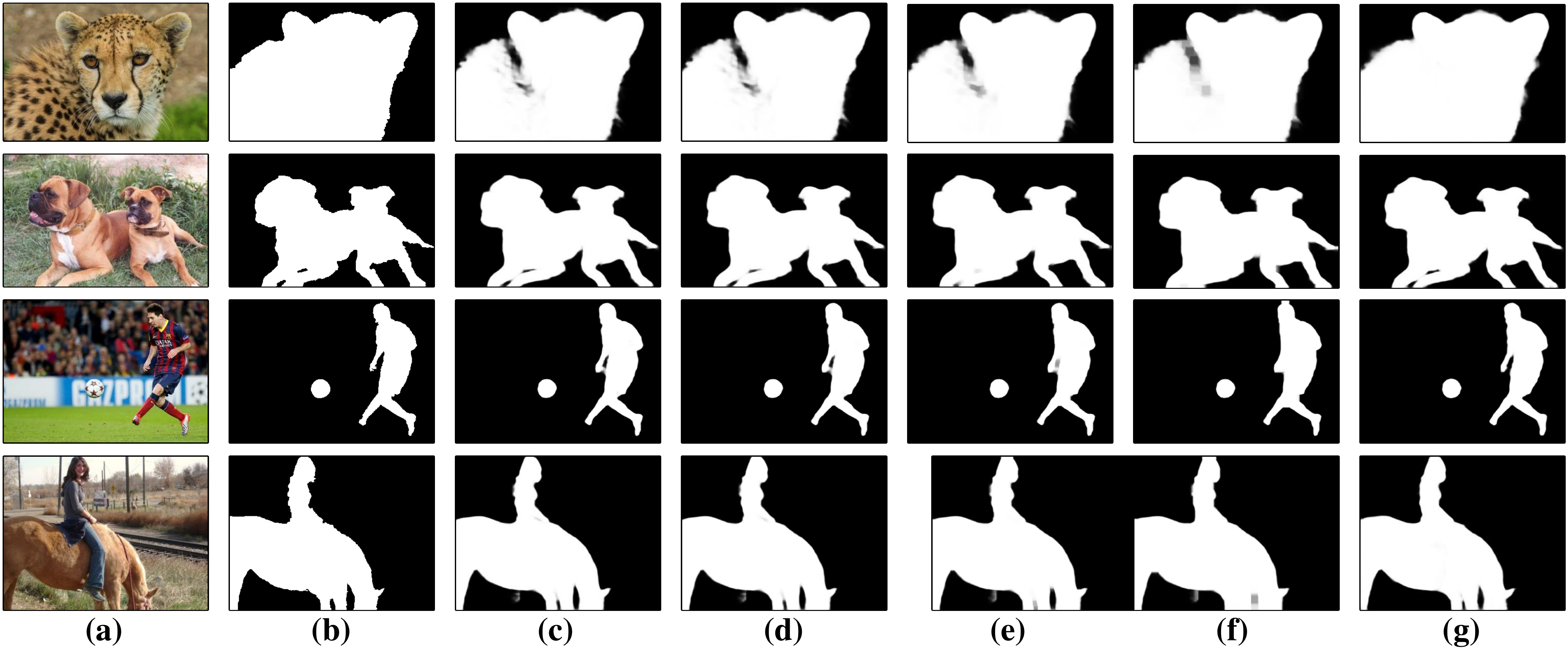}
% 	\vspace*{-6mm}
	\caption{Visual comparisons of our model with different kernel sizes of closing operation and PSG Loss (best viewed digitally with zoom). (a) Image, (b) GT, (c) Model4, (d)-(f) Model4 + $3 \times 3$, $7 \times 7$, and $13 \times 13$ closing, respectively. (g) Ours (Model4 + PSG Loss). Model4 denotes the fourth model variant in Table \ref{tab:ma}.}
	\label{fig:imagespgvp}
% 	\vspace{-6mm}
	% 	\vspace{-4mm}
\end{figure}

We also try to use the normal morphological closing operation as the post-processing method for refining the results of our models. We find that the evaluation performance of using a unified small kernel size of the closing operation, such as $3 \times 3$, $5 \times 5$, and $7 \times 7$, on all of the testing images, are roughly the same as the results of non-using. But it doesn't mean that the morphological closing post-processing is useless. In fact, there is an optimal kernel size for refining a specific image. This kernel size should be dependent on the size of its incomplete regions. As shown in the first row images in Fig. \ref{fig:imagespgvp}, using a small kernel of closing cannot complete the big 'holes' predicted by our Model4 model. For this type of image, a larger kernel is more suitable. For other images in Fig. \ref{fig:imagespgvp}, simply using a larger kernel of closing will worse the performance by wrongly merging the non-salient pixels into the salient regions. It usually occurs in images that contain some salient objects with clear but close boundaries. Moreover, closing operations cannot rectify the false positives. In this case, using a larger kernel of closing is also a bad choice. By comparing the results of using post-processing and those of using PSG loss, we can see that the accuracy of raw prediction is more important than the choice of the post-processing parameter.

To summarize, the performance of using the closing operation as post-processing is not as good as conducting it in the loss function. There are two main drawbacks to the former manner: firstly, the post-processing result is sensitive to the choice of kernel size. In PSG loss, a fixed and small kernel can be adopted as the model is progressively guided and the predicted result can be iteratively optimized through the epochs. Secondly, post-processing does not always yield better results. In PSG loss, correct auxiliary supervisions can be always guaranteed with the knowledge of $SM_{gt}$, which can help to prevent performance degradation.

\subsection{Application in Existing Methods}

Our proposed PSG loss is an auxiliary loss function that can be directly applied in training any end-to-end SOD models without any architecture modification. In this paper, we try to apply our PSG loss into three recent SOD methods: CPD \cite{wu2019cascaded}, PoolNet \cite{liu2019simple}, and  EGNet \cite{zhao2019egnet}, for evaluating the generalization ability of this loss. These models are retrained by using their released source codes as well as the default training settings. In Table \ref{tab:ma2}, we report the quantitative results of the original models and the models retrained with our PSG loss. We can see that PSG loss can further improve the performance of these existing models on almost all SOD benchmarks. 

\section{Conclusion}
\label{sec:con}

In this work, we have proposed a simple yet effective progressive self-guided (PSG) loss for assisting the training of deep learning-based salient object detection (SOD) models. Our PSG loss simulates the morphological closing operation on the model intermediate predictions for creating progressive and auxiliary training supervisions step-wisely. These progressively created training supervisions are always better than the current predictions, which can provide continuous and correct guidance to the SOD models. The effectiveness and the generalization ability of this loss have been validated in our experiments. Moreover, we also propose a new multi-scale feature aggregation module (MS-FAM), which is equipped with a branch-wise attention mechanism for further refining the learned features. Experimental results on six widely used SOD benchmark datasets have demonstrated the outstanding performance of our method with respect to other FCN-based models. In the future, we will consider promoting our PSG loss into other dense prediction tasks and investigating other post-processing techniques within this loss for further improvements.

\begin{figure}[!h]
	\centering
% 	\vspace*{-6mm}
	\includegraphics[page=8,trim = 5mm 5mm 5mm 5mm, clip, width=0.9\linewidth]{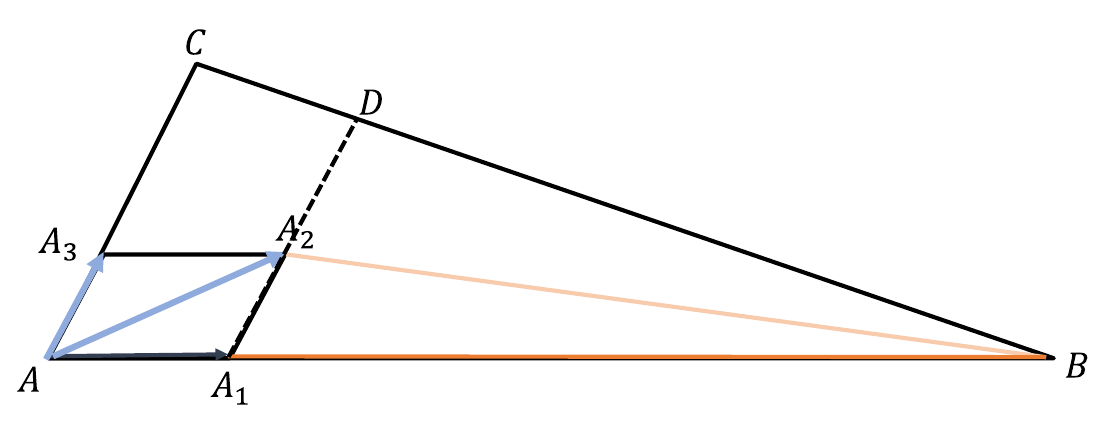}
% 	\vspace*{-6mm}
	\caption{An illustration of the relationship between the PSG loss and main loss. $A_1 B > A_2 B$ can be proved by using the geometric relationship. }
	\label{fig:images1hh}
% 	\vspace{-6mm}
	% 	\vspace{-4mm}
\end{figure}

\appendix 
In this appendix, we want to show that the optimized results by using our PSG loss with the main loss together can be better than the results by using the main loss only in a single ideal training step. Figure \ref{fig:images1hh} is used to illustrate and prove this. For simplicity, we use point $A, B, C$ to represent the current network prediction, $SM_{gt}$, and $SM_{pgt}$, respectively. Because our $SM_{pgt}$ is similar but more complete than the network predictions, $C$ is near to $A$ and the distance of $BC$ is smaller than the distance of $AB$. Besides, $A, B, C$ three points should be not collinear because the gradients from PSG loss are not proportional to those from the main loss. Since the loss function can encourage the network prediction to become more similar to its training supervision, the distance between them projected in this $ABC$ plane should be smaller. Here, we simply assume that the direction of gradient descent projected in this plane is the same as the direction of $\overrightarrow{AB}$ or $\overrightarrow{AC}$. In this ideal case, the main loss can move $A$ towards $B$ (assume it will move $A$ to $A_1$), and PSG loss can move $A$ towards $C$ (assume it will move $A$ to $A_3$). Therefore, using our PSG loss with the main loss together can move $A$ to $A_2$ according to the parallelogram law. We can use the geometric relationship to prove that the distance of $A_2 B$ is smaller than $A_1 B$, as presented in the following paragraph. It means that $A_2$ (using PSG loss with main loss together) is better than $A_1$ (using main loss alone) at this training step. Therefore, for each single ideal training step, our approach can get a better result with the help of PSG loss. As such, better performance can be progressively accumulated.

The proof of $A_1 B > A_2 B$: \par
We extend $A_1 A_2$ to cross BC in D, as depicted in Figure \ref{fig:images1hh}. \par

Because $AA_1 A_2 A_3$ is a parallelogram, we have \angle $CAB = $ \angle $A_2 A_1 B$ and  \angle $ACB = $ \angle $A_2 DB$.

Since $CB < AB$,  $\angle CAB < \angle ACB$, and hence $\angle A_2 A_1 B < \angle A_2 DB$.
Since $\angle A_1 A_2 B = \angle A_2 DB + \angle DBA_2 > \angle A_2 DB$, $\angle A_1 A_2 B > \angle A_2 A_1 B$. Therefore, we have  $A_1 B > A_2 B$.
% Since CB<AB, so ∠CAB< ∠ACB, and hence ∠A_2 A_1 B<∠A_2 DB

% \begin{proof}

% \end{proof}

\bibliographystyle{IEEEtran}
\balance
\bibliography{bibtex/ref_SOD}


% \end{IEEEbiography}


% % insert where needed to balance the two columns on the last page with
% %\newpage



%\vfill


\end{document}